\documentclass[lettersize,journal,twoside]{IEEEtran}
\usepackage{amsmath,amsfonts}
\usepackage{algorithmic}
\usepackage{algorithm}
\usepackage{array}
\usepackage[caption=false,font=normalsize,labelfont=sf,textfont=sf]{subfig}
\usepackage{textcomp}
\usepackage{stfloats}
\usepackage{url}
\usepackage{verbatim}
\usepackage{graphicx}
\usepackage{cite}
\usepackage{multirow}
\usepackage{tabularx}
\usepackage{hyperref}
\usepackage{threeparttable} 
\usepackage{amssymb}
\usepackage{xcolor}

\hypersetup{
	colorlinks=true,
	linkcolor=blue, 
	citecolor=blue, 
	urlcolor=blue, 
}

\begin{document}

\title{Grasp Like Humans: Learning Generalizable Multi-Fingered Grasping from Human Proprioceptive Sensorimotor Integration}

\author{Ce Guo$^*$, Xieyuanli Chen$^*$, Zhiwen Zeng, Zirui Guo, Yihong Li, Haoran Xiao, \\ Dewen Hu$^\dag$, Huimin Lu$^\dag$
	\thanks{This work is supported in part by the National Science Foundation of China under Grant U22A2059, 62203460, 62403478, and T2521006, Young Elite Scientists Sponsorship Program by CAST (No. 2023QNRC001), as well as the Innovation Science Foundation of National University of Defense Technology under Grant 24-ZZCX-GZZ-11.}
	\thanks{The authors are with the College of Intelligence Science and Technology, National University of Defense Technology. $^*$ indicates equal contributions.}
	\thanks{$^\dag$corresponding authors: Dewen Hu and Huimin Lu.}
}

\markboth{IEEE Transactions on Robotics}
{Guo \MakeLowercase{\textit{et al.}}: Grasp Like Humans: Learning Generalizable Multi-Fingered Grasping}


\maketitle
\begin{abstract}

Tactile and kinesthetic perceptions are crucial for human dexterous manipulation, enabling reliable grasping of objects via proprioceptive sensorimotor integration. For robotic hands, even though acquiring such tactile and kinesthetic feedback is feasible, establishing a direct mapping from this sensory feedback to motor actions remains challenging. In this paper, we propose a novel glove-mediated tactile-kinematic perception-prediction framework for grasp skill transfer from human intuitive and natural operation to robotic execution based on imitation learning, and its effectiveness is validated through generalized grasping tasks, including those involving deformable objects. 
Firstly, we integrate a data glove to capture tactile and kinesthetic data at the joint level. The glove is adaptable for both human and robotic hands, allowing data collection from natural human hand demonstrations across different scenarios. It ensures consistency in the raw data format, enabling evaluation of grasping for both human and robotic hands. Secondly, we establish a unified representation of multi-modal inputs based on graph structures with polar coordinates. We explicitly integrate the morphological differences into the designed representation,
enhancing the compatibility across different demonstrators and robotic hands. Furthermore, we introduce the Tactile-Kinesthetic Spatio-Temporal Graph Networks (TK-STGN), which leverage multidimensional subgraph convolutions and attention-based LSTM layers to extract spatio-temporal features from graph inputs to predict node-based states for each hand joint. These predictions are then mapped to final commands through a force-position hybrid mapping. 
Comparative experiments and ablation studies demonstrate that our approach surpasses other methods in grasp success rate, finger coordination, contact force management, and both grasp and computational efficiency, achieving results most akin to human grasping. The robustness of our approach is also validated through multiple randomized experimental setups, and its generalization capability is tested across diverse objects and robotic hands. The dataset and additional supporting videos are accessible via \url{https://grasplikehuman.github.io/}.

\end{abstract}


\section{Introduction}

\IEEEPARstart{B}{ionic} robotic hands have become a promising solution to many manipulation tasks~\cite{liang2021multifingered,hu2023grasping,andrychowicz2020learning,tian2022simplified}. Such hands surpass two-fingered grippers in providing enhanced dexterity to resemble human capabilities~\cite{liang2021multifingered}. However, achieving reliable finger coordination and contact force management for stable grasping remains a challenge, especially for objects with diverse properties, such as deformable ones. Drawing on visual feedback and prior knowledge, humans can pre-plan a grasping strategy tailored to an object's shape, hardness, and other properties. Even without vision, humans can still dynamically refine their grasp through touch, adjusting hand posture and contact forces in real-time. This reliability in grasping relies not only on manual dexterity and experience but also on the high-resolution tactile and kinesthetic feedback~\cite{yang2023tacgnn} and sensorimotor integration~\cite{wolpert1995internal,ordas2023neural}.
The latest definition describes sensorimotor integration as a process whereby sensory input is integrated by the central nervous system and used for assisting motor program execution~\cite{ordas2023neural}.

Inspired by human biological mechanisms, we aim to develop grasping skills for robotic hands by emulating human proprioceptive sensorimotor integration, i.e., the mapping from tactile and kinesthetic perception to hand joint movements. The goal is to enable bionic robotic hands to develop biomimetic grasping skills for handling objects of varying shapes, masses, and hardness, without relying on visual feedback or identifying each object individually. This requires effective force management and precise finger coordination to minimize excessive deformation and prevent object drops. As a result, our robotic hands can achieve reliable, adaptive grasps even without prior knowledge of the object, while avoiding potential damage caused by excessive forces.

For learning-based grasping approaches utilizing multimodal inputs, the collection of sufficient and accurate interaction data is critical. One common method for acquiring such data is to develop simulators that allow virtual hands to explore virtual environments~\cite{hu2023grasping,wan2023unidexgrasp++}. However, this approach presents challenges due to the complexity of simultaneously simulating robotic hands, manipulated objects with diverse properties, and their rich, nonlinear, and unpredictable interactions. Beyond creating highly detailed simulators, prior studies~\cite{zhao2023learning,funabashi2022multi,palleschi2023grasp} have demonstrated that learning from human demonstrations is an effective alternative for generating valuable imitation data.

Various methods have been proposed for obtaining human demonstrations, including three categories: passive observation, kinesthetic teaching, and teleoperation~\cite{ravichandar2020recent}. Passive observation~\cite{wang2023mimicplay,xu2022learning} appears straightforward for human demonstrators. However, challenges such as occlusion and interference from extraneous features frequently lead to issues like causal confusion~\cite{de2019causal}. Kinesthetic teaching~\cite{palleschi2023grasp,gabellieri2020grasp,wei2024wearable} and teleoperation~\cite{liang2021multifingered,zhao2023learning}
avoid modeling the mismatch between human and robot embodiments. However, these methods require demonstrators to consciously translate their intuitive actions into robotic equivalents, potentially resulting in incomplete representations of human ontological intelligence in the demonstration dataset. Moreover, the kinesthetic teaching and teleoperation demonstrations are linked to a specific manipulator, necessitating the recollection of all teaching data if the manipulator changes. To tackle this, we propose a scheme that directly captures human tactile and kinesthetic feedback along with instructional motions. Thus, the demonstrations can be applied to guide different robotic hands without demonstration recollection and model re-training.

Another key to realizing an imitation learning process based on tactile and kinesthetic feedback is establishing a unified representation of time-varying tactile and kinesthetic features~\cite{yang2023tacgnn}. In contrast to other modalities like images, tactile and kinesthetic features are characterized by their sparsity and topological correlation. Prior studies have employed Convolutional Neural Networks (CNNs) to encode tactile features~\cite{sundaram2019learning,funabashi2020stable}, necessitating the reorganization of the original feature to conform to the input specifications of CNNs. This restructuring process may lead to the loss of spatially correlated features within the dataset. To address this limitation, we employ Graph Convolutional Networks (GCNs) to simultaneously encode and interpret tactile and kinesthetic features, which operate directly on the connectivity relationships of graph nodes, effectively preserving spatial and topological correlations within the data.

In summary, we propose a learning-based approach imitated from human proprioceptive feedback to convey human manipulation skills to robotic hands. Its effectiveness has been validated through diverse grasping tests involving objects with challenging attributes, such as deformable, irregular, and slippery ones. This framework uses real-world interaction and motion data of human hands as templates, aiming to accurately capture and replicate their proprioception-based operations. The key contributions include:
\begin{itemize}
	\item[$\bullet$] We propose a novel glove-mediated tactile-kinematic perception-prediction framework, which enables manipulation skills across biological and artificial systems through standardized perceptual coding. The integrated data glove simultaneously acquires full-palm tactile and kinesthetic features at the joint level, enabling natural demonstrations by different demonstrators across different scenarios. 
	When worn by robotic hands, the glove provides real-time multi-modal feedback similar to that observed during human demonstrations. This consistent multi-modal feedback allows the trained model to adapt to new robotic hands via
	only simple manual calibration, avoiding the need for additional data
	collection or retraining. Furthermore, we establish unified evaluation metrics that comprehensively assess grasp performance across both human and robotic hands. 
	\item[$\bullet$] We propose a unified graph representation to encode joint motions and contact forces within a graph structure. Polar coordinates are used for such joint motions to emphasize key finger movement patterns. We then introduce novel Tactile-Kinesthetic Spatio-Temporal Graph Networks (TK-STGN) to predict desired joint motions and contact forces. This architecture leverages topological features through weighted multidimensional subgraph convolutions, enabling robust spatial feature extraction. Integrated LSTM layers with attention mechanisms further optimize temporal pattern analysis across motion sequences. 
	\item[$\bullet$] We establish a force-position mapping from desired joint features to robotic hand inputs. This mapping enables the generalization capability across different robotic hands. We validate this on six configurations of real robotic hands.
\end{itemize}

\section{Related Work}

\subsection{Robot Manipulation Based on Imitation Learning}
Using robotic hands in executing grasping deformable objects presents many challenges, such as predicting the multiple contact positions for reliable grasping, modeling nonlinear interactions, forecasting the desired contact force in the steady state, and the complexity of finger-gaiting~\cite{bicchi2000hands}. These difficulties hinder the use of modeling and optimization methods~\cite{sundaralingam2019relaxed,liu2022multi} and result in a significant performance gap when implementing identical learning-based approaches in simulated and physical environments~\cite{hu2021living,ding2021sim}. 
Collecting multimodal demonstration data by human demonstrators in physical environments has shown promise in tackling these challenges. 
For instance, kinesthetic teaching methods~\cite{palleschi2023grasp,gabellieri2020grasp} are used to obtain human grasping demonstration for standard cubes, thus enabling the grasping of unseen objects. 
Zhao et al.~\cite{zhao2023learning} present the ALOHA system that performs imitation learning directly from real demonstrations. They teleoperate using the leader robots, with the follower robots mirroring the motion, and eventually achieve dexterity operations with 80-90\% success. 
UMI~\cite{chi2024Universal} employs hand-held grippers to enable portable, low-cost, and information-rich data collection for challenging bimanual and dynamic manipulation demonstrations. Wei et al.~\cite{wei2024wearable} utilize kinesthetic teaching via a wearable robotic hand that simultaneously captures demonstrations and executes dexterous manipulation through hand-over-hand operation.

However, kinesthetic teaching~\cite{palleschi2023grasp,gabellieri2020grasp} and teleoperation methods~\cite{liang2021multifingered,zhao2023learning} require the operator to perform tasks from the robot's perspective, which is unintuitive and misaligned with natural human operation. UMI~\cite{chi2024Universal} and HIRO Hand~\cite{wei2024wearable} enable demonstrators to perform operations from their perspectives. Nevertheless, the data collection process still relies on specific teaching devices, which constrains the replication of human hand dexterity and the imitation of human proprioceptive intelligence. Furthermore, these device-mediated imitation learning methods face difficulties in replacing the manipulators, which can lead to incompatibility or necessitate the recollection of demonstration data.

\subsection{Tactile Sensing and Tactile-based Manipulation}

Many of the studies on dexterous manipulation rely on visual inputs~\cite{wan2023unidexgrasp++,xu2023unidexgrasp,wang2023mimicplay,wei2024learning}. However, visual feedback faces challenges such as occlusion, inability to capture proprioception, and the lack of direct force observation~\cite{yang2023tacgnn}. In the presence of nonlinear elastic deformations or plastic deformations in the manipulated objects, tactile feedback becomes crucial.
Some related studies focus on integrating visuo-tactile technology~\cite{zhao2023fingerslam,yan2022detection,wang2020swingbot} or novel material techniques~\cite{yao2022encoding,lin2021skin} to design precise tactile sensing devices. They are used for tasks such as object localization and reconstruction~\cite{zhao2023fingerslam}, slip detection~\cite{yan2022detection}, dynamic swing-up~\cite{wang2020swingbot}, etc.

However, these sensors are often configured only at the fingertips of robotic hands due to structural complexity or expensive price~\cite{lepora2021towards, murali2018learning,garg2019learning,zhao2024tactile}.
In this work, we argue that the distribution and topology of tactile sensing also matter, especially for contact-rich tasks. 
For example, Sundaram et al.~\cite{sundaram2019learning} used a scalable tactile glove to show that sensors uniformly distributed over the hand can identify individual objects, estimate their weight, and explore the typical tactile patterns that emerge while grasping objects.

\begin{figure*}[!t]
	\centering
	\includegraphics[width=7.0in]{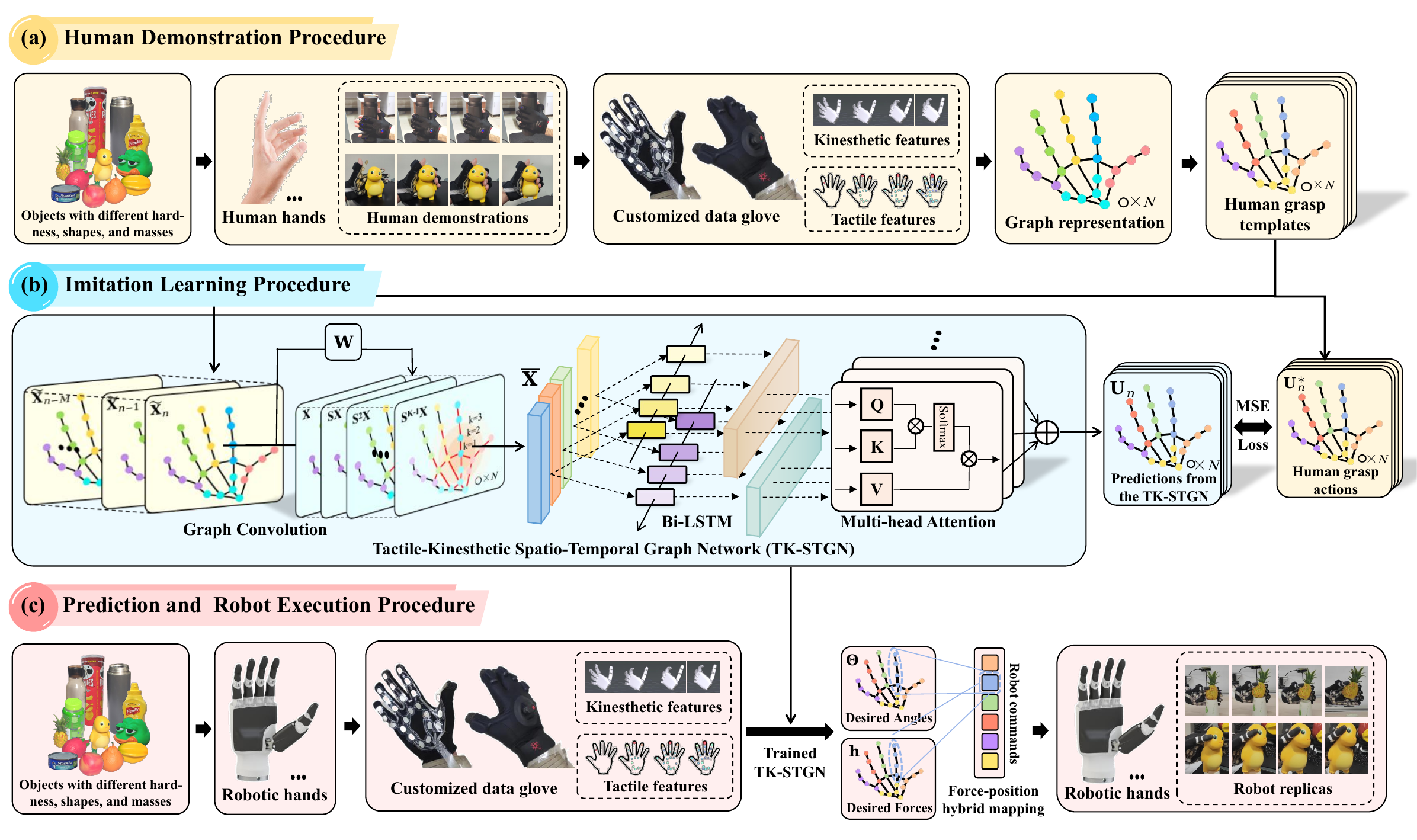}%
	\caption{Pipeline of our framework. (a) In the demonstration procedure, the human operator wears the data glove to complete the grasping operation, and the tactile and kinesthetic features are integrated as graph data, which is sampled to become an expert template. (b) In the imitation procedure, the graph data is fed into the TK-STGN, and the mean square errors of the TK-STGN's outputs with respect to the expert templates are computed to optimize the parameters. (c) In the prediction and execution procedure, the robotic hand wears the data glove to perform grasping operations. Its tactile and kinesthetic perceptions are transformed into graph data and input into the trained TK-STGN. The desired node-based state output from the network generates the action commands through force-position hybrid mapping to regulate the robot's real-time movements.}
	\label{fig_1:Overview of our approach}
	\vspace{-8pt}
\end{figure*}

\subsection{Data Glove with Embedded Tactile-Kinesthetic Feedback}

Data gloves are widely employed across various domains, including hand pose reconstruction~\cite{park2024stretchable}, object identification and property inference~\cite{sundaram2019learning}, and immersive teleoperation with force-motion feedback~\cite{wang2024tactile,shen2023fluid}.
For instance, triboelectric sensing with pneumatic actuation on the data glove enables comprehensive tactile-kinesthetic sensing and feedback during virtual object interactions, replicating texture and mechanical resistance properties~\cite{wang2024tactile}. Additionally, stretchable liquid-metal sensors integrated into the data glove enable precise hand pose reconstruction for cross-platform teleoperation via anatomically conformal motion capture~\cite{park2024stretchable}.
Advanced tactile and kinesthetic sensing technologies~\cite{yao2022encoding,park2024stretchable}, coupled with the miniaturization and modularization trends of commercial sensors, have accelerated the development of integrated tactile-kinesthetic perception data gloves. This progress enables glove-mediated natural demonstration-to-execution frameworks for generating cross-platform dexterous manipulation strategies.

\subsection{GNNs for In-hand Manipulation}
When employing images to represent tactile~\cite{zhao2023fingerslam}, or tactile sampling points are dense~\cite{sundaram2019learning}, CNNs are viable solutions for encoding tactile features. However, in many cases, the tactile sampling points are sparse~\cite{yan2022robotic,yan2023geometric}, leading to sparse and non-Euclidean tactile feedback. In such cases, employing CNNs requires rearranging the feedback to match the desired rectangular input.
To preserve the topological characteristics inherent in sparse tactile features, certain studies represent these features as graphs. Consequently, graph neural networks (GNNs) are leveraged as a toolkit for analyzing and learning from graph-structured data~\cite{garcia2019tactilegcn,funabashi2022multi,yang2023tacgnn,dreher2019learning}.
Despite abundant studies employing GNNs on dexterous manipulation~\cite{funabashi2022multi,yang2023tacgnn,dreher2019learning}, the integration of multimodal inputs within graph structures remains an open issue.

\section{METHODOLOGY}

In this section, we describe the main components of our approach. Fig.~\ref{fig_1:Overview of our approach} shows the pipeline, including the human demonstration procedure, the imitation learning procedure, and the prediction and execution procedure. Specifically, tactile and kinesthetic feedback is captured through the data glove and is uniformly encoded as graph data, which is later fed into the TK-STGN. TK-STGN outputs joint state predictions that ultimately guide the motion of the robotic hands. We present a detailed description of the proposed approach's components as follows.

\subsection{Acquisition of Tactile and Kinesthetic Feedback}

\begin{figure}[!t]
	\centering
	\includegraphics[width=3.4in]{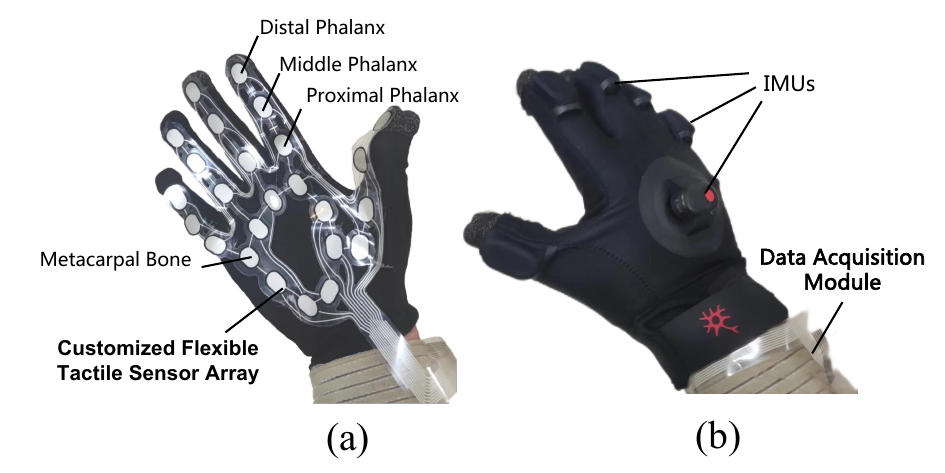}%
	\caption{The data glove. (a) The palm side of the glove. It is covered by the customized flexible tactile sensor array. (b) The back side of the glove. There are six IMUs spread out here.}
	\label{fig_2:the_data_glove}
	\vspace{-8pt}
\end{figure}

To obtain training data directly from human operations rather than from operating robotic hands~\cite{liang2021multifingered,funabashi2022multi}, we have integrated a data glove capable of capturing tactile feedback and joint motions while allowing for intuitive and natural operation by humans. This data glove enables the establishment of the glove-mediated tactile-kinematic perception-prediction framework for transferring human proprioception-based skills to various robotic hands.

The palm of the data glove is covered with 25 tactile sampling pads, located on the contact surface of the distal phalanx, middle phalanx, proximal phalanx, metacarpal bone of each finger, and the lower part of the palm, as shown in Fig.~\ref{fig_2:the_data_glove}(a). The flexible tactile sensor array is created by applying force-sensitive materials, such as sensitive ink and silver paste, onto a flexible film substrate using a precise printing process, followed by drying and curing\footnote{\url{http://www.roxifsr.com/}}. The hollowed-out portion of the substrate, located between the fingers and at the palmar center, is designed to enhance mechanical flexibility. The tactile pads on the array are capable of a maximum positive pressure of approximately 20\,N and a sampling frequency exceeding 200\,Hz. Each tactile pad of the data gloves is calibrated using standard weights. In this paper, we treat the acquired magnitude and distribution of contact forces as a downscaled but reasonable observation of tactile features~\cite{tian2023tactile,li2024easycalib,liu2022multi,lepora2021towards}.
The back of the glove integrates the PN3 Pro, an inertial motion capture device developed by Noitom\footnote{\url{https://www.noitom.com/}}, as depicted in Fig.~\ref{fig_2:the_data_glove}(b). This device can capture the Euler angles in the local coordinate system and the position and orientation in the global coordinate system of the 20 hand joints at a frequency of approximately 70\,Hz. The IMU corresponding to each finger should be located between the middle and proximal phalanges. Calibration via flexion and opposition movement is needed before each use.

By wearing the data glove, the demonstrator can demonstrate directly with their own hands, ensuring a natural and intuitive demonstration process. The glove captures real-time tactile and kinesthetic feedback with minimal interference to the operation. Additionally, the glove can be worn directly or with minor modifications on a humanoid robotic hand, acquiring multi-modal features in the same format as the human hand. The data glove, including its data acquisition module, has a total weight of 65.2 g and can work continuously for over four hours. The PN3 Pro offers skeletal length templates tailored to different genders and heights and allows for the input of customized skeletal length templates. This feature enhances the precision in capturing the kinematic characteristics of various demonstrators' hands and robotic hands.

\begin{figure}[!t]
	\includegraphics[width=2.7in]{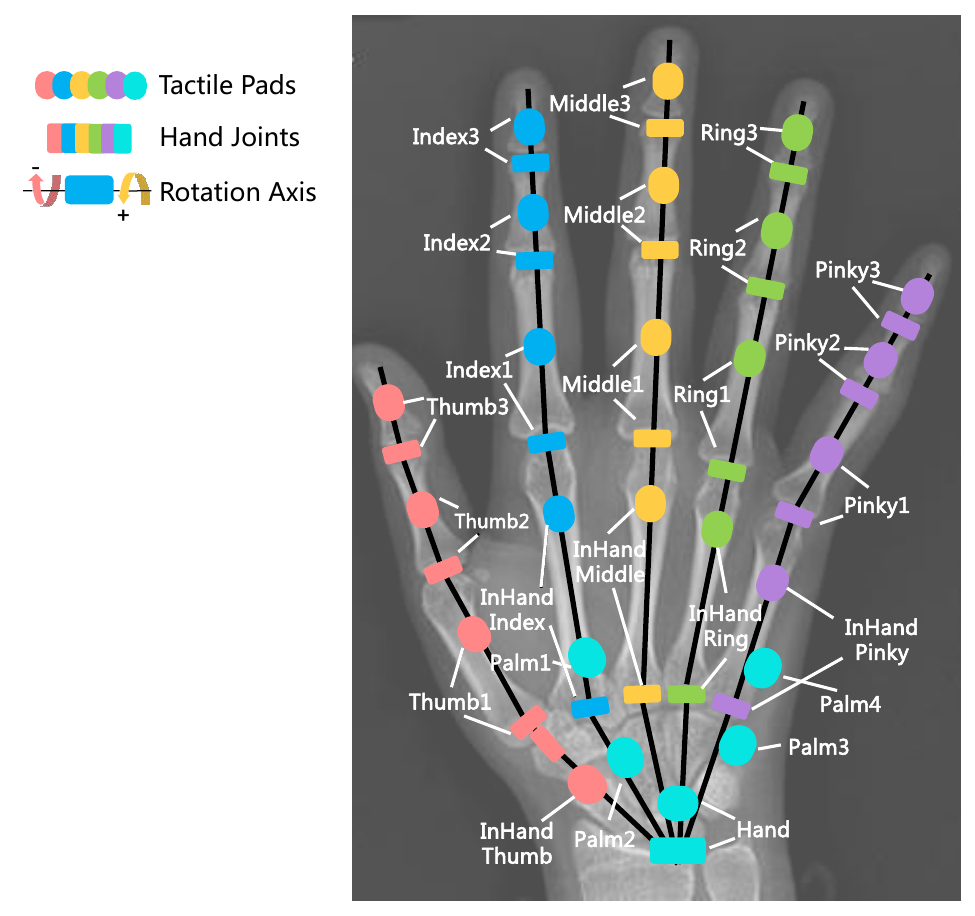}%
	\caption{The distribution of 25 tactile sampling pads and 20 hand joints. Among the 25 tactile pads, 19 are positioned to directly correspond to the major bones of the human hand, including the phalanges and metacarpal bones. Each of these 19 pads is associated with the joints that primarily determine its local posture, such as interphalangeal joints and metacarpophalangeal joints, and is labeled identically for a further unified encoding of the multi-inputs. The remaining six sampling units (Palm1-Palm4, InHandThumb, and Hand) are evenly distributed across the lower region of the palm. Since the intermetacarpal joints exhibit minimal relative motion~\cite{standring2005gray}, which cannot be effectively measured, five of these units (Palm1-Palm4 and InHandThumb) are not directly linked to any specific joint. The Hand pad, however, is specifically bound to the motion of the carpometacarpal joint. This figure presents a view from the palm side of the left hand.}
	\label{fig_3:tactile_pads_hand_joints}
	\vspace{-8pt}
\end{figure}

Fig.~\ref{fig_3:tactile_pads_hand_joints} shows the distribution and labeling of 25 tactile sampling pads and 20 hand joints. The corresponding joint and tactile pad are labeled identically for a further unified encoding of the multi-inputs. 

In summary, based on the data glove, we ensure the consistency of data formats when used by different human and robotic hands. This consistency guarantees uniform input features for both the training and prediction procedures, eliminating the need for remapping caused by perceptual absence or structural differences in robotic hands. In addition, based on the uniform features captured by the data glove, we further establish unified evaluation metrics for both human and robotic hand operations, as detailed in Sec.~\ref{sec:force_eval_metrics}.

\subsection{Unified Representation of Multi-modal Inputs}\label{sec:unified_repre}

\begin{figure}[!t]
	\centering
	{\includegraphics[width=2.5in]{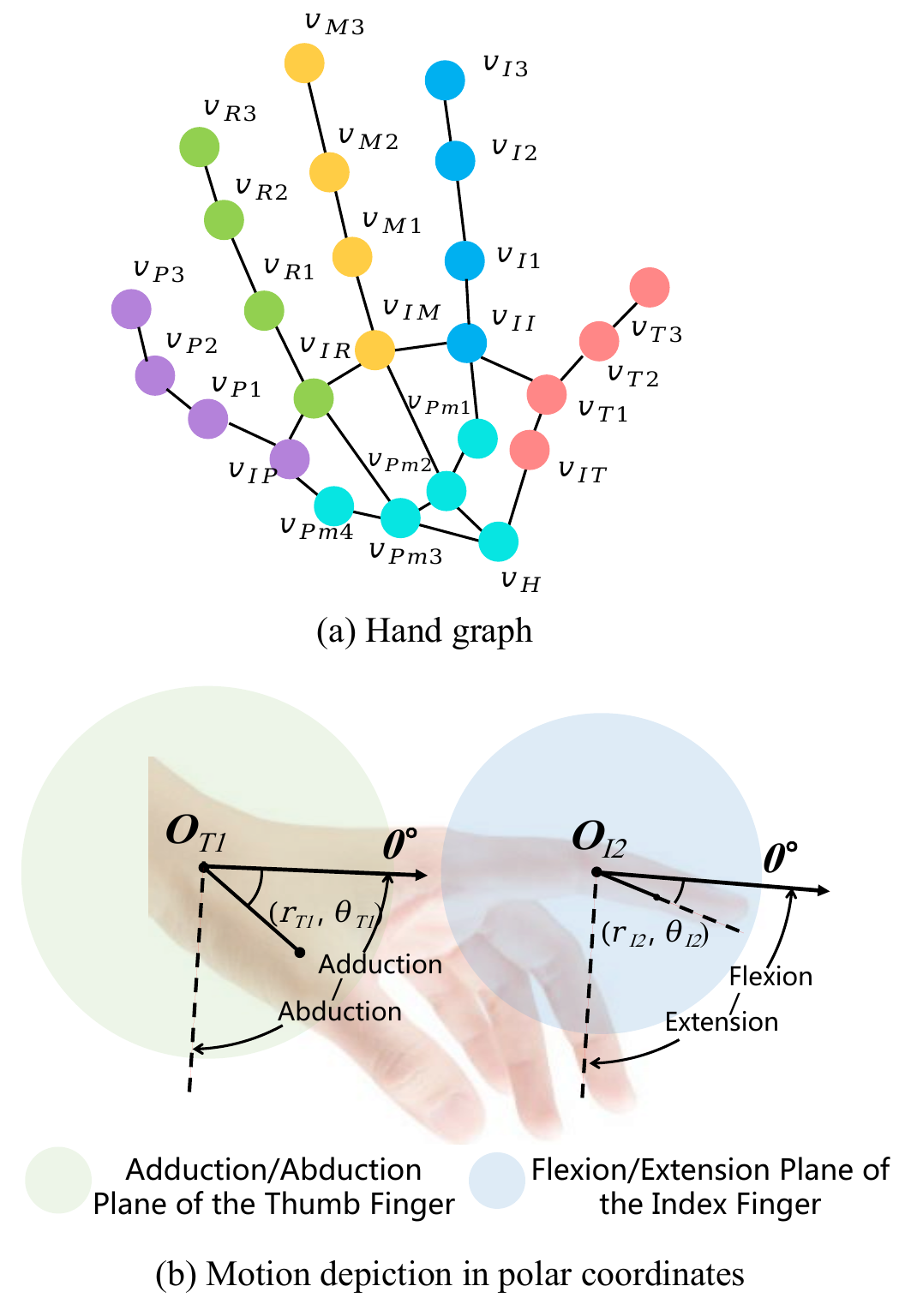}%
		\label{fig_4:graph}}
	\caption{Unified representation based on the graph and polar coordinates. (a) The hand graph. The edges in the palm are simplified for aesthetic purposes. (b) The motions' depiction of the thumb metacarpal and the index finger's middle phalanx within their respective polar coordinate systems. $r$ and $\theta$ represent the bone length and joint angle, respectively.}
	\label{fig_4:unified_encode}
\end{figure}

Compared to common visual feedback, the joint motion features and tactile features acquired in this study are sparse and exhibit significant topological correlations. Utilizing image-like encoding requires a large amount of data reordering, affecting the efficiency and effectiveness of extracting structure-related features. Therefore, we adopt the use of graphs to organize multimodal data. Fig.~\ref{fig_4:unified_encode}(a) illustrates the hand graph $\mathcal{G}=(\mathcal{V},\mathcal{E})$, where the node set $\mathcal{V}=(v_{T1},v_{T2},\cdots,v_{Pm4},v_{H})$ represents hand joints and corresponding tactile pads shown in Fig.~\ref{fig_3:tactile_pads_hand_joints}. The subscripts of the nodes are abbreviations of the labels used in Fig.~\ref{fig_3:tactile_pads_hand_joints}. The edge set $\mathcal{E}$ indicates the structural correlation between them. The structural characteristics of the hand, including the relative independence of finger proprioception and movement, as well as the dense distribution of muscles between the metacarpals~\cite{standring2005gray}, suggest that proprioceptive coupling is stronger in the palm. Therefore, nodes on the fingers are only connected sequentially to neighboring finger nodes, while nodes in the palm are fully connected. The nodes directly connected to $v_i$ are called neighbors of $v_i$, denoted $\mathcal{N}(v_i)$. Nodes that require traversing at least $k$ edges to establish interaction are defined as $k$-hop neighbors of $v_i$. The subgraph containing all the $\tilde{k}$-hop neighbors $(\tilde{k}=0,1,2,\cdots,k)$ and the edges between them are called the $k$-hop subgraph of $v_i$.

Based on the hand graph, the multi-modal hand data can be uniformly represented as state vectors bound to the nodes:
\begin{equation}
\mathbf{x}_i \triangleq \begin{bmatrix} \mathbf{x}_{ki}  \\
\mathbf{x}_{ti} \end{bmatrix},
\end{equation}
where $\mathbf{x}_{ki}$ and $\mathbf{x}_{ti}$ represents the kinesthetic and tactile feedback of $v_i$, respectively. The $v_{PmX}$ nodes correspond to the lower region of the palm, where the metacarpal bones are located. These nodes only provide tactile feedback because the morphological changes in this part of the palm are difficult to capture. According to~\cite{standring2005gray}, the metacarpal bones have highly intricate structures, yet their relative displacement is negligible. To ensure feature alignment, we assign the kinematic feedback of the carpometacarpal joints to the $v_{PmX}$ nodes.

We describe hand motion within the joint space because it offers a more concise and effective representation of movement compared to Cartesian space~\cite{liang2021multifingered,liu2022multi}. 
Besides, to achieve a comprehensive understanding of motion dynamics across various demonstrators, we incorporate the factor of bone length into our analysis. By combining this factor with the identification of motion patterns in joint
space, we create a representation of joint motion in polar coordinates. Specifically, by establishing a polar coordinate system with the joint position as the origin of the principal motion plane, the bone lengths and joint angles are effectively translated into coordinates within the polar framework.
Fig.~\ref{fig_4:unified_encode}(b) illustrates the motion of the thumb metacarpal and the middle phalanx of the index finger within their respective polar coordinate systems as examples. These coordinate systems correspond to the abduction/adduction of the thumb and the flexion/extension of the index finger. As illustrated in Fig.~\ref{fig_4:unified_encode}, the polar coordinates-based motion representation selectively captures key finger movements essential for grasping~\cite{standring2005gray}, specifically finger flexion/extension and thumb abduction/adduction. This avoidance of over-parameterized motion representations enables skill transfer with fewer demonstrations while enhancing operational robustness.
In addition, angular velocity is considered another important reference for recognizing human hand motion characteristics and controlling joint movements. Thus, the kinesthetic feedback of $v_i$ can be expressed as:
\begin{equation}
\mathbf{x}_{ki} \triangleq [r_i,\theta_{i},\dot{\theta}_{i}]^\top,
\end{equation}
where $r_i,\theta_i$, and $\dot{\theta}_{i}$ represent the bone length, joint angle, and angular velocity of $v_i$, respectively. They can be captured in real-time through the inertial motion capture device.

In summary, the multi-modal state vector of $v_i$ at $n$-th sampling moments are defined as:
\begin{equation}
\mathbf{x}_{i,n} \triangleq [r_i,\theta_{i,n},\dot{\theta}_{i,n},h_{i,n}]^\top,
\end{equation}
where $h$ represents the magnitude of contact force captured through the tactile sensor array.

\begin{figure}[!t]
	\centering
	\includegraphics[width=3.3in]{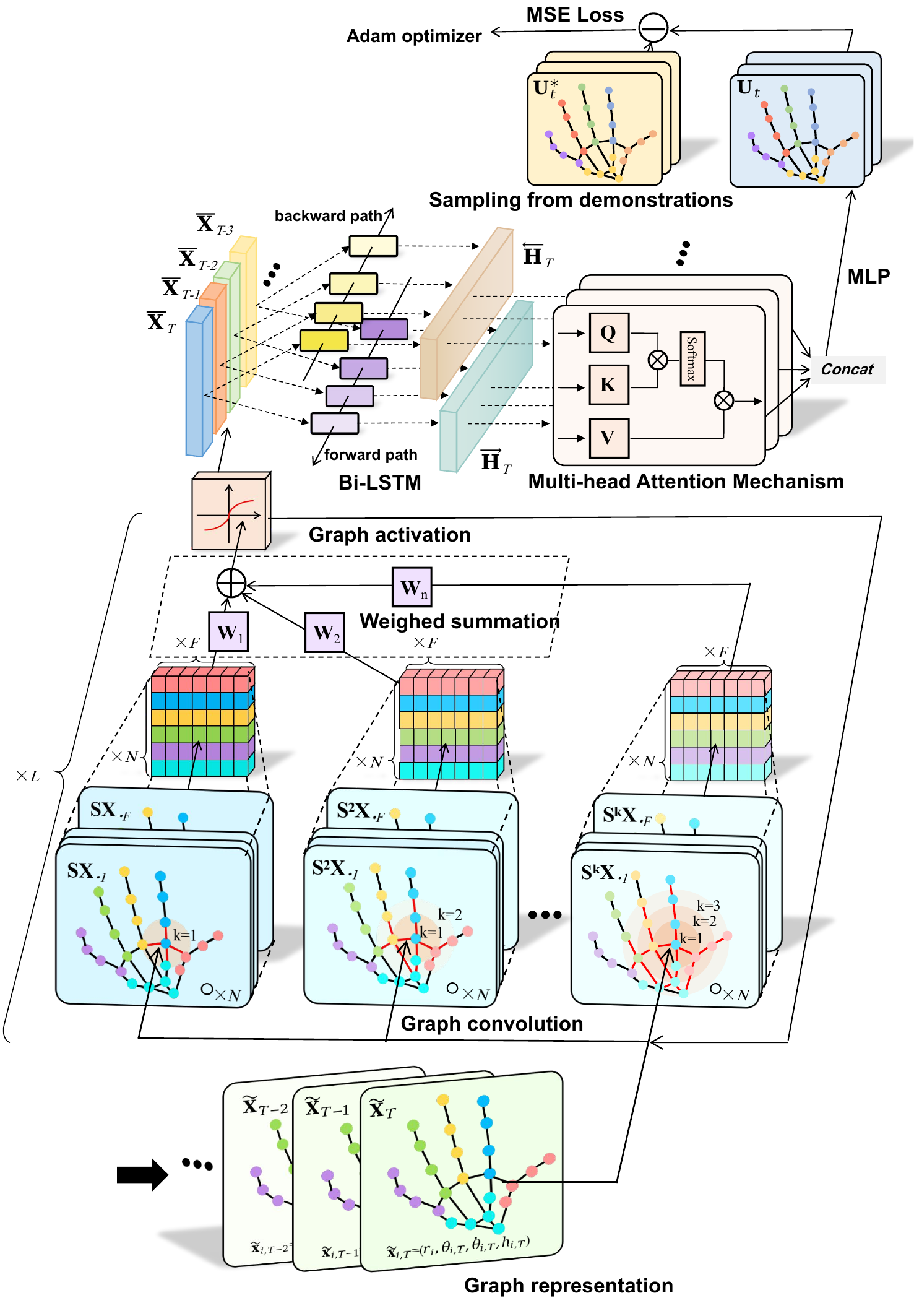}%
	\caption{The structure of TK-STGN and imitation learning procedure. The input to TK-STGN consists of $M$ historical states, including node-structured kinesthetic and tactile features. TK-STGN comprises $L$ sequentially connected graph convolutional layers followed by temporal feature extractors. Specifically, the graph convolutional layers perform linear combinations of graph shift operations on multi-dimensional subgraphs, while the graph activation layers enhance the network's ability for nonlinear mapping. Spatial patterns captured by graph convolutions are then fed into bidirectional LSTM modules to model temporal dynamics, with subsequent multi-head attention mechanisms highlighting task-critical features across time steps. This hierarchical architecture enables robust spatiotemporal feature encoding for grasp perception. During training, the MSE loss is computed between the predicted states generated by TK-STGN and the target states sampled from the human demonstration training dataset. The parameters are then optimized using the Adam optimizer.}
	\label{fig_5:The structure of our graph convolutional networks.}
	\vspace{-10pt}
\end{figure}

\subsection{Imitation Learning Based on TK-STGN}\label{sec:GCNs}

Once obtaining the node-based multi-modal inputs $\mathbf{X} =[\mathbf{x}_1, \mathbf{x}_2, \cdots, \mathbf{x}_N]^\top $ data from our glove, we introduce Tactile-Kinesthetic Spatio-Temporal Graph Network (TK-STGN) as the backbone to extract features from the grasping state spatially and temporally. The structure of TK-STGN and imitation learning procedure are depicted in Fig.~\ref{fig_5:The structure of our graph convolutional networks.}. 

The graph convolution operation can be defined as a weighted linear combination of graph shift operations in multidimensional subgraphs:
\begin{equation}
\mathcal{H}_n(\mathbf{X}_n,\mathbf{S}_n) = \sum_{k=0}^{K-1}\mathbf{S}_n^{k}\mathbf{X}_n\mathbf{W}_k,
\end{equation}
where $\mathbf{S}_n \in \mathbb{R}^{N \times N}$ is the $n$-th graph shift operator, which describes the relevance of the graph nodes at $n$-th sampling moment. $N$ is the total number of nodes in the hand graph. $\mathbf{X}_n \in \mathbb{R}^{N \times F}$ represents the $n$-th observation matrix of the hand nodes. $F$ is the dimensionality of the observations. $\mathbf{W}_k \in \mathbb{R}^{F \times D} $ represent the filter coefficient matrix to be optimized, which is shared across all nodes. Here, $K$ specifies the maximum order of the subgraph, defining the maximum range of the subgraph from which a node aggregates information for its output computation.

In this paper, we regard the symmetrically normalized adjacency matrix as the graph shift operator to reduce feature scale variations. Furthermore, we assume that the topological relationships of nodes remain unchanged during the grasping process. Therefore, $\mathbf{S}_n$ can be expressed as:
\begin{equation}
	\mathbf{S}_n = \mathbf{D}^{-\frac{1}{2}}\mathbf{A}\mathbf{D}^{-\frac{1}{2}},
\end{equation}
where $\mathbf{D} \in \mathbb{R}^{N \times N}$ is the degree matrix, and $\mathbf{A} \in \mathbb{R}^{N \times N}$ is the adjacency matrix.

The TK-STGN comprises $L$ sequentially connected graph convolutional layers, each containing a graph convolutional layer and a graph activation layer. The graph activation layer can be expressed by:
\begin{equation}
	\mathbf{X}_{l+1} = \sigma(\mathcal{H}_l(\mathbf{X}_l,\mathbf{S})),
\end{equation}
where $\mathbf{X}_{l+1} \in \mathbb{R}^{N \times F_{l+1}},\mathbf{X}_{l} \in \mathbb{R}^{N \times F_{l}}$ represent the output and input state matrices of $l$-th layer, respectively, and $\sigma$ denotes the graph activation function.

Spatial patterns captured by graph convolutions are then fed into bidirectional LSTM modules to model temporal dynamics, which can be expressed by:
\begin{align}
(\overrightarrow{\mathbf{h}}_n,\overrightarrow{\mathbf{c}}_n) &= \textit{LSTM}\left( \mathbf{\bar{X}}_n, (\overrightarrow{\mathbf{h}}_{n-1},\overrightarrow{\mathbf{c}}_{n-1}) \right), \\
(\overleftarrow{\mathbf{h}}_n,\overleftarrow{\mathbf{c}}_n) &= \textit{LSTM}\left(  \mathbf{\bar{X}}_n, (\overleftarrow{\textbf{h}}_{n+1},\overleftarrow{\mathbf{c}}_{n+1}) \right),
\end{align}
\begin{align}
\mathbf{H}_T &= \left[\overrightarrow{\mathbf{H}}_T \parallel \overleftarrow{\mathbf{H}}_T \right] \\
& = \left[ (\overrightarrow{\mathbf{h}}_1, \overrightarrow{\mathbf{h}}_2, \dots, \overrightarrow{\mathbf{h}}_T) \parallel ( \overleftarrow{\mathbf{h}}_1, \overleftarrow{\mathbf{h}}_2, \dots, \overleftarrow{\mathbf{\mathbf{h}}}_T) \right],
\end{align}
where $\parallel$ denotes feature concatenation, $\overrightarrow{\mathbf{h}}_n $ and $\overleftarrow{\mathbf{h}}_n$ denote the forward and backward hidden states at timestep $n$, $\overrightarrow{\mathbf{c}}_n$ and $\overleftarrow{\mathbf{c}}_n$ represent the corresponding cell states capturing long-term temporal dependencies.
$\overrightarrow{\mathbf{H}}_T$ and $\overleftarrow{\mathbf{H}}_T$ 
denote the tensors concatenating all forward and backward states over $T$ timesteps. This bidirectional processing captures both causal and anticausal dependencies in the grasping process, which is essential for anticipating contact transitions during dynamic grasping.

Subsequently, a $M$-head attention mechanism prioritizes task-critical features across temporal windows:

\begin{align}
\mathbf{a}_m &= \textit{Softmax}\left(\frac{\left(\mathbf{H}_T \mathbf{W}_{Q,m}\right)\left(\mathbf{H}_T \mathbf{W}_{K,m}\right)^\top}{\sqrt{d_k}}\right)\left(\mathbf{H}_T \mathbf{W}_{V,m}\right),\\
\mathbf{H}_{\text{att}} &= [\mathbf{a}_1 \parallel \mathbf{a}_2 \parallel \cdots \parallel \mathbf{a}_M] \mathbf{W}_O,
\end{align}
where $d_k$ represents the dimension of the key and query vectors per attention head, $\mathbf{W}_{Q,m},\mathbf{W}_{K,m},\mathbf{W}_{V,m}$ are learnable projection matrices for queries, keys and values of the $m$-th head, $\mathbf{a}_m$ is the context vector from the $m$-th attention head, and $\mathbf{W}_O$ is the output projection matrix.

The tensor $\mathbf{H}_{\text{att}}$ is then passed through a multi-layer perceptron (MLP) to generate the future predictions for $P$ steps:
\begin{equation}
	\mathbf{\hat{U}} = \textit{MLP}(\mathbf{H}_{\text{att}}),
\end{equation}
where $\mathbf{\hat{U}}$ contains predicted joint angles and contact forces.

Unlike other imitation learning processes that take the robot action commands as the target outputs, we aim to predict the desired states that match the human hand's joint topology through imitation to ensure generalization capability among different robotic hands. Specifically, the output of TK-STGN at $n$-th moment can be expressed as:
\begin{align}
	\mathbf{\hat{U}_n} &= [\mathbf{\hat{u}}_{1,n},\mathbf{\hat{u}}_{2,n},\cdots,\mathbf{\hat{u}}_{N,n}]^\top,\\
	\mathbf{\hat{u}}_{i,n} &\triangleq [\hat{\theta}_{i,n+1},\hat{h}_{i,n+1},\hat{b}_{i,n}]^\top,
\end{align}
where $\hat{\theta}_{i,n+1}$ and $\hat{h}_{i,n+1}$ represent the predicted joint angle and contact force of $v_i$ for the next moment, respectively, and $\hat{b}_{i,n}$ represents the determination of whether the current state is a steady state. The corresponding target output from the templates is expressed as:
\begin{align}
\mathbf{U}_n^* & = [\mathbf{u}^*_{1,n},\mathbf{u}^*_{2,n},\cdots,\mathbf{u}^*_{N,n}]^\top,\\
\mathbf{u}^*_{i,n} & \triangleq [\theta^*_{i,n+1},h^*_{i,n+1},b^*_{i,n}]^\top,
\end{align}
where $\theta^*_{i,n+1}$ and $h^*_{i,n+1}$ are collected by the data glove and $b^*_{i,n}$ is manually annotated, and $\mathbf{U}_n^* \in \mathbb{R}^{N \times 3}$ represents the target outputs sampled from the demonstration dataset.

The input and output structure of TK-STGN aligns with the perception input and state output at the joint level of human hands. Combined with the use of our data glove, this design enables the collection of demonstration data for the same task in different scenarios and from various demonstrators. Furthermore, replacing different robotic hands does not affect the state prediction performance of TK-STGN, eliminating the need to recollect demonstration data or retrain the model when a new robotic hand is introduced.

The objective of our imitation learning procedure can be expressed as:
\begin{equation}
\mathbf{W}^* = \arg\min_{\mathbf{W}} \sum_{\mathcal{T}_n \in \mathcal{T}} \mathcal{L}(\mathbf{\hat{U}}_n,\mathbf{U}_n^*), 
\end{equation}
where $\mathbf{W}^*$ represents the optimal coefficient matrix, $\mathcal{T}$ represents the demonstration dataset, and $\mathcal{L}$ is the loss function computed by weighted Mean Square Error (MSE). The loss $\mathcal{L}$ can be expressed as:
\begin{equation}
	\mathcal{L} = \sum_{p=1}^P \omega_p 
	\left(
	\frac{1}{N} \sum_{i=1}^N \left[ (\hat{\theta}_{i,p} - \theta_{i,p}^*)^2 + (\hat{h}_{i,p} - h_{i,p}^*)^2 \right]
	\right),
\end{equation}
\begin{equation}
	\omega_p = \frac{e^{-p} }{ \sum_{p=1}^P e^{-p}}
\end{equation}
where $\omega_p$ is the weight for step $p$ and exponentially decaying with $p$, and $P$ is the prediction horizon.

In summary, TK-STGN enables node-based joint state prediction at the human joint level, replicating the multi-joint control strategies of the human hand under proprioceptive feedback while providing real-time evaluation of grasp reliability. Utilizing the consistency of multimodal representations, any robotic hand equipped with the data glove can acquire real-time desired joint angle and contact force references, which reflect the human hand's dexterity and stability criteria.

\subsection{Force-Position Hybrid Mapping}

In Sec.~\ref{sec:GCNs}, we obtain the desired joint node state prediction $\mathbf{U}_n$ that matches the dimensions of the human hand's action space. To translate $\mathbf{\hat{U}}_n$ into action commands for the robotic hand, we develop a hybrid mapping method that combines force and position. The mapping scheme can be expressed as:
\begin{equation}
	\mathbf{c}_n = \mathbf{\Gamma}_\theta (\mathbf{\hat{\Theta}}_{n+1} - \mathbf{\Theta}_n)+ \mathbf{\Gamma}_h (\mathbf{\hat{h}}_{n+1} - \mathbf{h}_n),
\end{equation}
with vector definitions:
\begin{align*}
	\mathbf{\hat{\Theta}}_{n+1} &= [\hat{\theta}_{1,n+1}, \hat{\theta}_{2,n+1}, \cdots, \hat{\theta}_{N,n+1}]^\top,\\
	\mathbf{\Theta}_n &= [\theta_{1,n}, \theta_{2,n}, \cdots, \theta_{N,n}]^\top,\\
	\mathbf{\hat{h}}_{n+1} &= [\hat{h}_{1,n+1}, \hat{h}_{2,n+1}, \cdots, \hat{h}_{N,n+1}]^\top,\\
	\mathbf{h}_n &= [h_{1,n}, h_{2,n}, \cdots, h_{N,n}]^\top,
\end{align*}
where $\mathbf{c}_n \in \mathbb{R}^{C}$ represents the action commands, $C$ is the active degrees of freedom, $\mathbf{\hat{\Theta}}_{n+1}, \mathbf{\hat{h}}_{n+1} \in \mathbb{R}^{N}$ represent the predicted joint angles and contact forces, $\mathbf{\Theta}_n, \mathbf{h}_n \in \mathbb{R}^{N}$ represent the joint states captured by the data glove,  and $\mathbf{\Gamma}_\theta,\mathbf{\Gamma}_h\in \mathbb{R}^{C\times N}$ are the sparse matrices containing the coefficients determined experimentally.

$\mathbf{\Gamma}_\theta$ and $\mathbf{\Gamma}_h$ represent the correlation between joint angles and contact force errors with the commands. The correlation coefficients are assessed based on various factors, including the robotic hand's actuated degrees of freedom and its morphological similarities to the human hand. In the case of underactuated robotic hands, the command input for each actuator is related to the angle and contact force deviations of multiple associated joints. For example, the flexion or extension of the index finger should be determined collectively by Index1-3, InHandIndex, Palm1, and Palm2 nodes, which are located in or near the flexion/extension plane of the index finger. In contrast, $\mathbf{\Gamma}_\theta$ and $\mathbf{\Gamma}_h$ are diagonal for fully actuated robotic hands, meaning that the command input for each joint actuator is correlated only with the angle and contact force error of its corresponding joint.
Note that $\mathbf{\Gamma}_\theta$ and $\mathbf{\Gamma}_h$ remain unchanged while dealing with different objects, unless the robotic hand is replaced. 

In summary, our approach generates suitable operation commands for both underactuated and fully actuated robotic hands, as long as the dimensionality of their action space does not exceed that of a human hand. By leveraging dimensionality reduction mapping, this method eliminates the need to collect additional demonstration data or retrain the model when deploying on different robotic hands.

Building on all the above, our framework enables continuous integration of demonstration data from various demonstrators into the expert dataset, which enhances the robustness of the generated grasping skill. Furthermore, it facilitates the generalization of the same skill to robotic hands with differing configurations. 

\section{Dataset Collection and Network Training}

\begin{figure}[!t]
	\centering
	\includegraphics[width=3.5in]{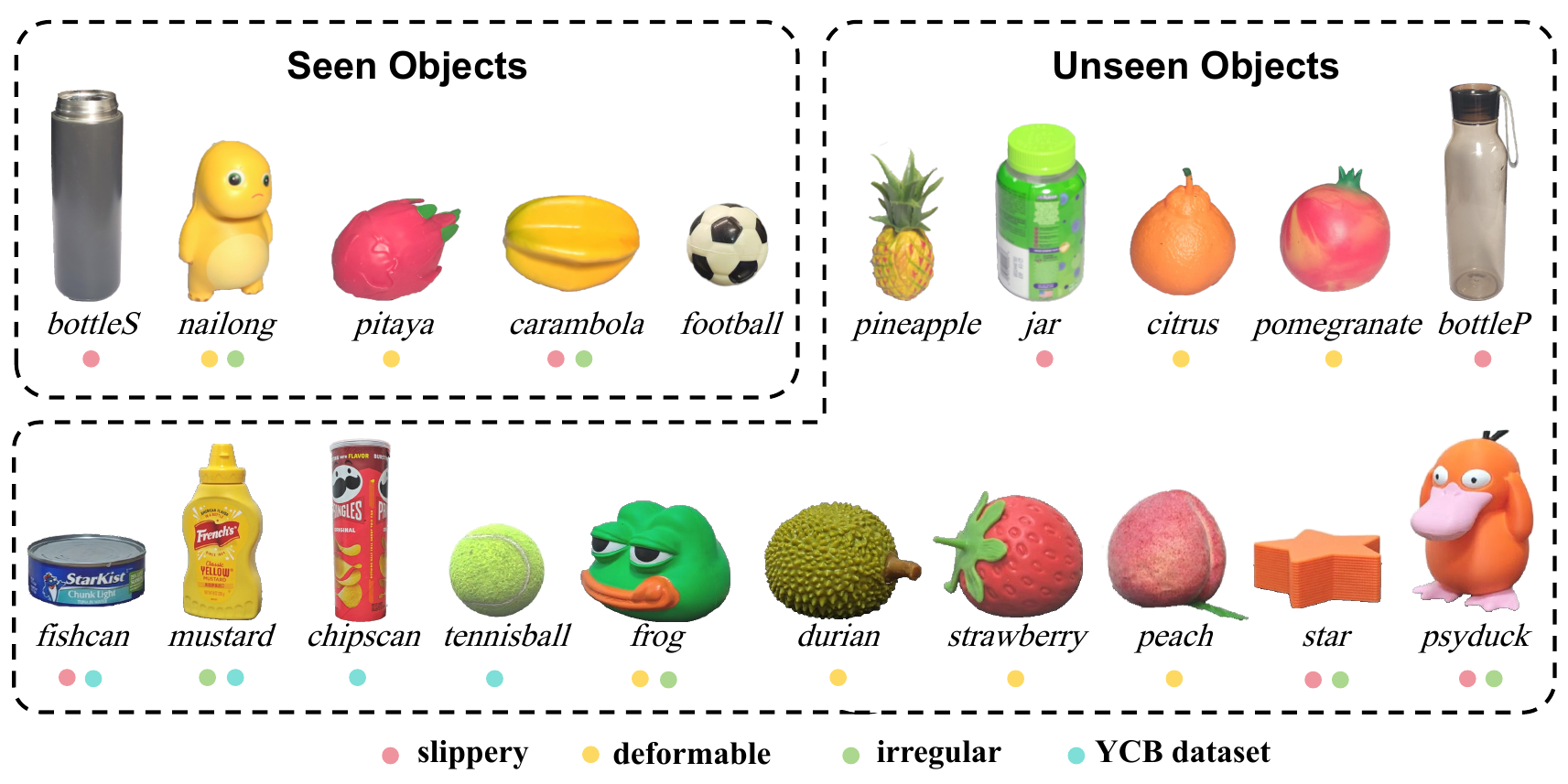}%
	\caption{Seen and unseen objects used in the training and testing process.}
	\label{fig_7:objects_in_datasets}
	\vspace{-8pt}
\end{figure}

This section details the collection methodology for demonstration data, including visualization protocols for tactile and kinesthetic features during grasp execution. The experimental dataset encompasses both seen objects used during training and unseen objects for generalization testing, as shown in Fig.~\ref{fig_7:objects_in_datasets} and quantitatively characterized in Tab.~\ref{tab_1:objects}.

\begin{figure}[!t]
	\centering
	{\includegraphics[width=3.4in]{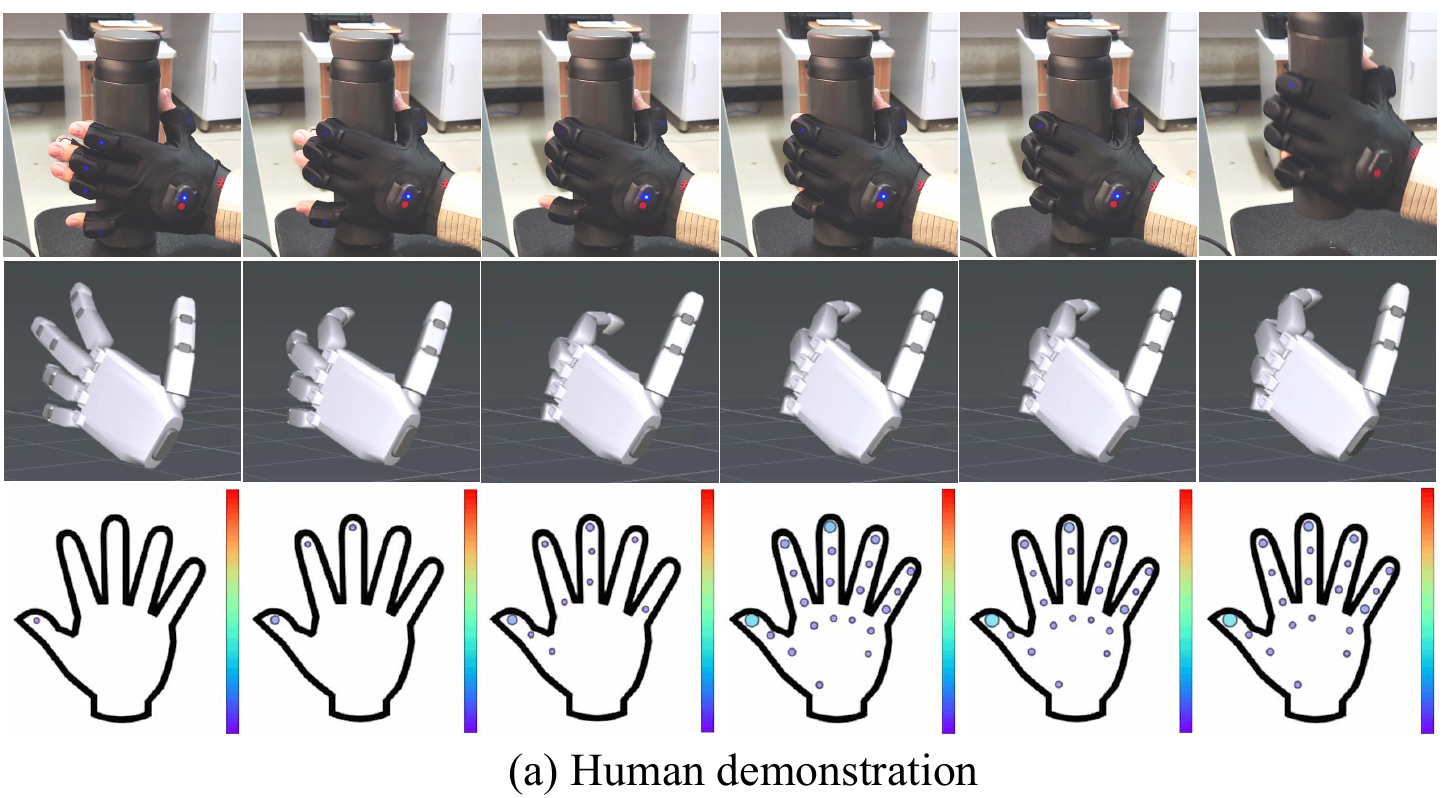}%
		\label{fig_6a:human_grasp_demo}
		\hfil
		\includegraphics[width=3.4in]{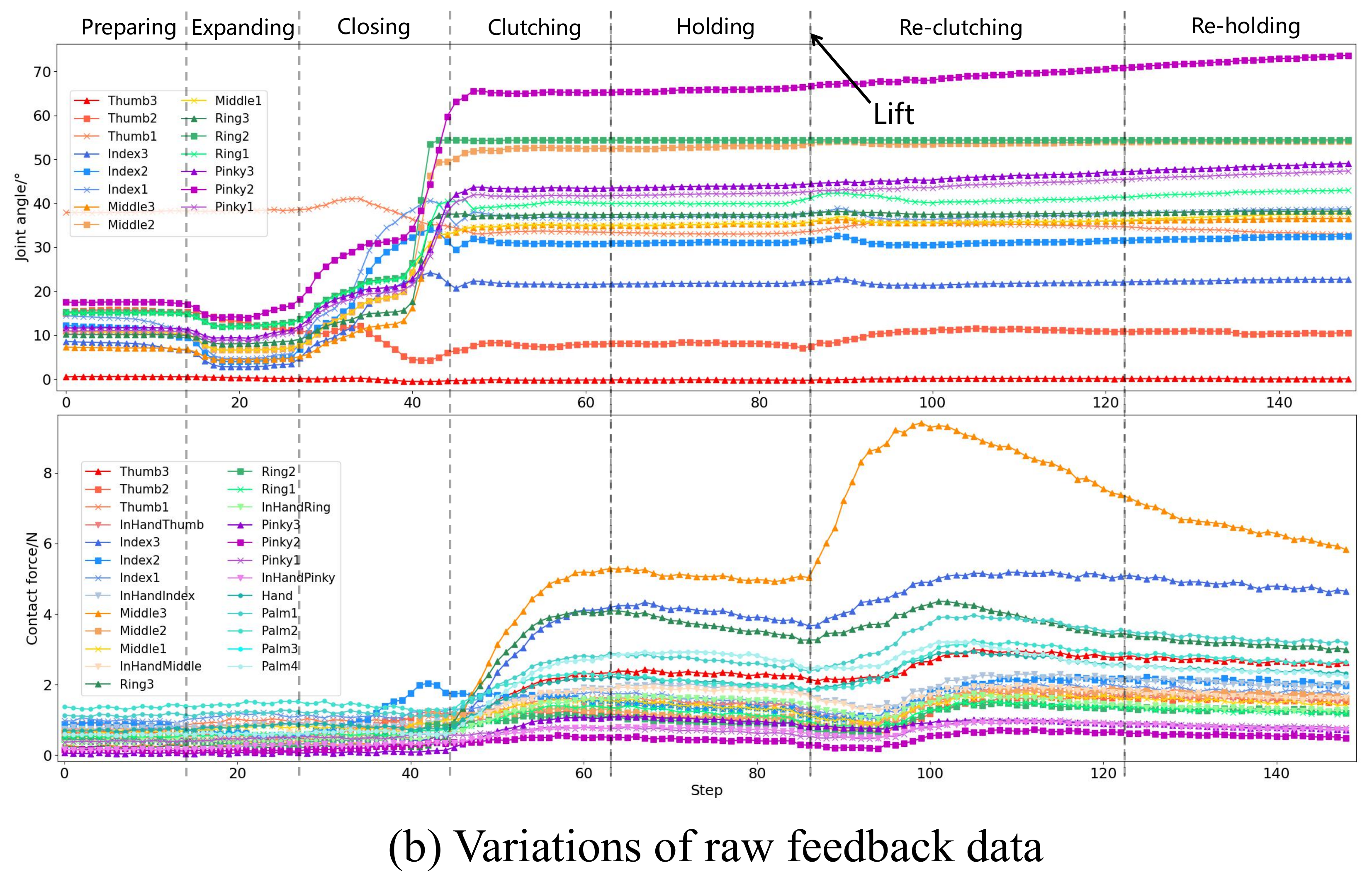}%
		\label{fig_6b:human_grasp_angle_tactile}}
	\caption{A demonstration of grasping by a human hand. (a) Human grasping process, hand posture reconstruction, and visualization of contact force magnitude and distribution. (b) The variations of main joint angles and contact force magnitude. Data from different joints of each finger are labeled with similar colors. Triangles, rectangles, forks, inverted triangles, and dots are used to label the distal phalange, middle phalange, proximal phalange, and metacarpal bone of each finger and other sampling points in the palm, respectively.}
	\label{fig_6:human_grasp}
	\vspace{-8pt}
\end{figure}

The data glove help us obtain rich human proprioceptive sensorimotor integration features to build a demonstration dataset.
Fig.~\ref{fig_6:human_grasp}(a) shows the process of grasping and lifting the bottle. We realize the reconstruction of hand posture and the visual display of the contact force magnitude and distribution based on the acquired kinematic and tactile features. As demonstrated in the tactile feature visualizations, contact force magnitudes are encoded through proportional point scaling and color gradients, with the visual mapping calibrated to represent the range from 0 to approximately 20~N.
Fig.~\ref{fig_6:human_grasp}(b) shows the variations of main joint angles and contact force magnitude during the grasping process. Except for Thumb1, the joint angles represent Euler angles within the flexion/extension plane of the fingers. For the thumb, the flexion/extension motion mainly occurs at the Thumb2 and Thumb3 joints, while the abduction/adduction motion is mainly reflected in the feedback of Thumb1~\cite{standring2005gray}. Therefore, the value corresponding to Thumb1 is set as the abduction angle. All joint angles are zero when the fingers are fully extended.

\begin{table}[!t]
	\caption{The properties of the grasped objects. \label{tab_1:objects}}
	\centering
		\begin{tabularx}{0.495\textwidth}{m{0.66cm}<{\centering}m{0.65cm}<{\centering}m{0.74cm}<{\centering}m{1.36cm}<{\centering}m{0.86cm}<{\centering}m{1.07cm}<{\centering}m{0.57cm}<{\centering}}
			\hline
			\textbf{Label}     &\textbf{Mass} &\textbf{Hardn-ess}    &\textbf{Shape(Size)}         &\textbf{Mate-rial} &\textbf{Elastic Modulus} &\textbf{Grasp Mode} \\ \hline
			\textit{bottleS} &201.1g  & $>$90HA & cylindrical (65*194mm)    &stainless steel  &196GPa  & side       \\ 
			\textit{nailong}   & 48.6g   & 15HC   & subcylindrical (70*115mm) & TPE &0.204GPa  & side       \\
			\textit{pitaya}    & 185.5g  & 32HC   & spheroidal (83mm)   & TPE    & 0.204GPa   &top-down         \\
			\textit{caram-bola} & 18.7g   & 39-50HA & subcylindrical (66*103mm)   &MDPE    & 0.648GPa    &top-down         \\
			\textit{footb-all}    & 10.7g  & 16HC   & spheroidal (58mm)   & TPE    & 0.204GPa  &top-down         \\
			\textit{pinea-pple}   & 26.9g  & 75-93HA   & subcylindrical (70*133mm)  & MDPE    & 0.648GPa  &side         \\
			\textit{jar}    & 49.1g  & $>$90HA   & cylindrical (68*118mm)   & PET         & 3.25GPa    &side    \\
			\textit{citrus} & 167.4g  & 31HC   & spheroidal (78mm)
			& TPE         & 0.204GPa  &side         \\
			\textit{pomeg-rante}  & 133.8g  & 32HC   & spheroidal (74mm)   & TPE         & 0.204GPa  &top-down         \\
			\textit{bottleP}  & 103.9g  & $>$90HA   & cylindrical (60*228mm)   & PC  &2.6GPa   &side         \\
			\textit{fishcan}  & 175.6g  & $>$90HA   & cylindrical (83*33mm)   & stainless steel  &196Gpa   &top-down         \\
			\textit{mustard}  & 259.5g  & 66-87HA   & subcylindrical (72*155mm)   & HDPE  &0.995GPa   &side         \\
			\textit{chipscan}  & 194.8g  & 73-87HA   & cylindrical (78*232mm)   & Kraft Paper  &130MPa   &side         \\
			\textit{tennis-ball}  & 54.5g  & 65HA  & spheroidal (66mm)   & Butyl Rubber  &7.77GPa   &top-down        \\
			\textit{frog}  & 76g  & 14-26HC   & subcylindrical (109*81mm)   & TPE  &0.204GPa   &top-down        \\
			\textit{durian}  & 159.4g  & 13-20HC   & spheroidal (97mm)   & TPE  &0.204GPa   &top-down         \\
			\textit{straw-berry}  & 171.9g  & 20-23HC   & spheroidal (69mm)   & TPE  &0.204GPa	   &top-down         \\	
			\textit{peach}  & 8.8g  & 16HA   & spheroidal (84mm)   & MDPE  &0.648GPa   &side         \\
			\textit{star}  & 20.8g  & 84-96HA   & subcylindrical (90*30mm)   & PLA  &5.92GPa   &top-down         \\
			\textit{psyduck}  & 56.9g  & $>$90HA   & subcylindrical (92*88mm)   & PLA  &5.92GPa   &side         \\
			\hline
	\end{tabularx}
\end{table}

The distinct phases of human hand grasping, including preparing, expanding, closing, clutching, holding, re-clutching, and re-holding, can be clearly identified through data variations. Joint angle changes mainly characterize the process from the closing phase to just before contact. In contrast, during the clutching phase, joint angles stabilize after contact, while the contact force gradually increases and stabilizes. The demonstrator lifts the bottle between the holding and re-clutching phases, causing a noticeable change in the contact force and leading to a new force equilibrium. 
In summary, both kinesthetic and tactile feedback reflect the human grasping experience, and they are important inputs for realizing stable grasping.

Fig.~\ref{fig_7:objects_in_datasets} presents the twenty selected objects exhibiting diverse physical properties, including mass, hardness, shape, and elastic modulus. Five seen objects are used to build the demonstration dataset, while the remaining fifteen unseen objects are used for testing the generalization ability of different approaches. Key properties are annotated in this figure, including deformable, slippery, and irregular. To improve the standardization of the test, we introduce some objects from the YCB dataset~\cite{calli2015benchmarking} as unseen objects.

The properties of the grasped objects are listed in Tab.~\ref{tab_1:objects}. Hardness is measured using Shore A and Shore C durometers, and the values for the elastic modulus are obtained as the average of the material properties referenced from a material database MatWeb\footnote{\url{https://matweb.com/index.aspx}}. 
We use Shore durometers with an accurate measurement range between 10 and 90\,HA/HC. For objects with hardness values exceeding 90\,HC, the Shore A durometer is used, while objects with values below 10\,HA are measured using the Shore C durometer. We regard objects with a hardness exceeding 90\,HA as rigid bodies that cannot be deformed during grasping. For certain objects, hardness varies across different sites, resulting in a range of hardness values.

\begin{table}[!t]
	\caption{Training Hyperparameters and Setup\label{tab_2:Hyper}}
	\centering
	\begin{tabularx}{0.3\textwidth}{m{2.8cm}<{\centering}m{2cm}<{\centering}}
		\hline
		\textbf{Parameter} & \textbf{Value} \\   	\hline
		$N$ & 25\\
		$L$ & 2\\
		$K$ & 3\\
		$M$ & 4\\
		Optimizer & Adam \\  
		Learning Rate & $1e^{-4}$ \\ 
		Batch Size & 32 \\
		\hline
	\end{tabularx}
	\vspace{-8pt}
\end{table}

The demonstrators conduct 30 grasps for each object, with each session lasting 10 seconds and data sampled at 15\,Hz. During the first 5 seconds, the demonstrator transitions from an extended hand position to a stable grasp. In the remaining 5 seconds, the object is lifted to a specified height while maintaining a stable grasp. The demonstration data is sourced from both the left and right hands, as well as from different demonstrators. See Sec.~\ref{sec:Training with Expanding Datasets} for more details.

TK-STGN is built using Pytorch and runs on Ubuntu 20.04 with an NVIDIA GeForce RTX 3060 Ti GPU. Tab.~\ref{tab_2:Hyper} summarizes the key hyperparameters and training setup details for TK-STGN.

\section{Experiments}

In this section, we introduce novel metrics for grasp performance evaluation, detailed in Sec.~\ref{sec:force_eval_metrics}. Beyond success rates, these metrics emphasize force management during grasping, as well as grasping and computational efficiency. Leveraging the generalization capabilities of our data glove, these metrics enable direct comparisons of grasping performance across different robotic hands and human hands. We also provide extensive comparative experiments and ablation studies in Sec.~\ref{sec:Comparative Evaluation and Ablation Experiments}, demonstrating that our approach outperforms existing methods in grasp success rate, finger coordination, grasp force management, and both grasp and computational efficiency. The results indicate that our approach achieves performance closely resembling human grasping. The robustness and generalization capabilities of our approach are also validated in Sec.~\ref{sec:Robustness and Generalization Evaluation}. More experimental details are also provided in the supplementary video.

\subsection{Experimental Bionic Robotic Hand Setups}

\begin{figure}[!t]
	\centering
	\includegraphics[width=3.0in]{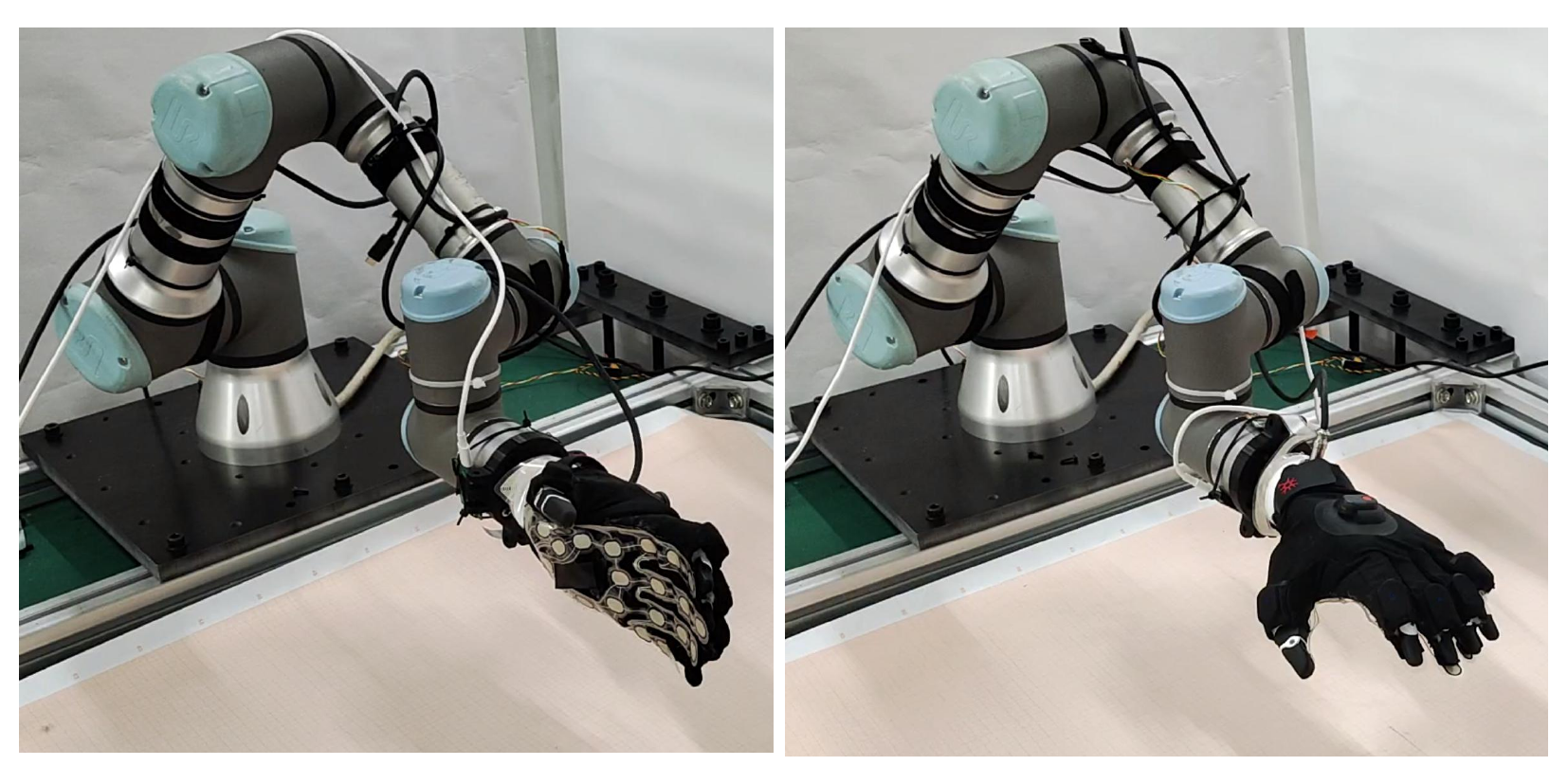}%
	\caption{The grasp testing system. The left and right figures show the initial positions of the side grasp and the top-down grasp, respectively.}
	\label{fig_8:grasp testing systems}
	\vspace{-8pt}
\end{figure}

As illustrated in Fig.~\ref{fig_8:grasp testing systems}, the grasp testing system is constructed using a UR3 robotic arm\footnote{\url{https://www.universal-robots.com/}} and an Inspire robotic hand\footnote{\url{https://en.inspire-robots.com/}}. Our data glove is mounted on the robotic hand, directly outputting multimodal perceptual features that are isomorphic to human demonstrations. In Sec.~\ref{sec:Comparative Evaluation and Ablation Experiments}, we assume that the robotic hand has been guided to a suitable grasping position, with the hand's initial posture fixed and the object's relative position to the hand fixed. In Sec.~\ref{sec:Robustness and Generalization Evaluation}, we test the performance for random invalid tactile pads, random approach angles, random initial postures, and random placement positions. The PN3 Pro and tactile pads are calibrated before use.

\subsection{Metrics for Grasp Performance Evaluation}
\label{sec:force_eval_metrics}

We believe that the ideal grasping approaches should dynamically adjust motions to achieve the appropriate force balance for each object,
which provides sufficient friction to counteract gravity while preventing excessive deformation or potential damage to the object. Thus, we present several grasp evaluation metrics, including success rate, force evaluation metrics (FEMs), and some temporal efficiency evaluation metrics. 

\textbf{SR} (Success Rate) is obtained based on 15 attempts for each object. A successful attempt means that the robotic hand is able to grasp the object and lift it to a specified height without dropping it during the sampling time.

The evaluation of contact force is implemented based on the tactile feedback from our data glove. Assessing contact force is inherently complex due to its simultaneous involvement in the object (20 objects $\times$ 15 attempts), period (10 seconds), and height (25 sampling positions). Consequently, the force evaluation is divided into three distinct metrics. Additionally, to eliminate the influence of failed grasps on force evaluation, we randomly selected five successful grasps from fifteen attempts for each object. If fewer than five successful grasps are achieved, additional trials are conducted to meet the requirement.

\textbf{FEM-AT} (All forces at the Terminal stage) is designed to assess the magnitude and consistency of all contact forces at the terminal stage of grasping. Given the independence of tactile feedback across different objects and sampling points, we aggregate the mean and standard deviation of contact forces from various objects and sampling points to serve as a comprehensive metric for the force evaluation. FEM-AT calculates the cumulative mean $F_{AT\_aver}$ and cumulative standard deviation $F_{AT\_std}$ by:
\begin{equation}
F_{AT\_aver} = \sum_{o=1}^{O} \sum_{i=1}^{N} \frac{\sum_{j=1}^{S} h_{t\_oij}}{S},
\end{equation}
\begin{equation}
F_{AT\_std} = \sum_{o=1}^{O} \sum_{i=1}^{N} \sqrt{\frac{\sum_{j=1}^{S}(h_{t\_oij} - \bar{h}_{t\_oi})^2}{S}},
\end{equation}
where $h_{t\_oij}$ represents the $i$-th contact force at the terminal stage of grasping for the $j$-th successful attempt of the $o$-th object, and $\bar{h}_{t\_oi}$ is the mean contact force at the terminal stage of grasping for the $i$-th contact force across $S$ successful attempts of the $o$-th object. Here, $O$ represents the number of objects grasped during the testing process, $N$ is the number of force sampling pads, and $S$ represents the number of successful grasps considered per object in the force evaluation. 
In this study, the evaluation focuses on five successful grasps for each of the 20 objects, with data collected at 25 sampling pads. Consequently, the parameters are set as follows: $O=20$, $N=25$, and $S=5$.

\textbf{FEM-KE} (Key forces during the Entire process) is designed to evaluate the magnitude of the top five key forces throughout the entire grasping process. In our experiments, we find that there are always several critical contact pads that provide the main contact force during each grasp. The key forces are selected by sorting the average value of contact forces across all selected attempts for a given object. The value of FEM-KE, denoted as $F_{KE}$, is defined as:
\begin{equation}
	F_{KE} = \frac{1}{5} \sum_{t=1}^{5} \left( \frac{1}{O S T} \sum_{o=1}^{O} \sum_{j=1}^{S} \sum_{n=1}^{T} h_{oi_njn} \right),
\end{equation}
where $h_{oi_njn}$ represents the $i_n$-th contact force at the $n$-th time for the $j$-th successful attempt of the $o$-th object. The indices $i_n$ correspond to the top five sampling points with the highest mean forces across all selected grasps. $T$ denotes the number of sampling times per grasp. With a sampling frequency of 15\,Hz over 10 seconds, $T$ is set to 150.

\textbf{FEM-MAE} (Maximum for All forces during the Entire process) is used to measure the instantaneous force performance throughout the entire grasping process, and it is determined by the maximum contact force observed across all selected grasping attempts and sampling pads.

At equivalent success rates, we believe that approaches with lower FEM values better control force during grasping, avoiding excessive force and reducing the risk of object damage.

We additionally identify two metrics to evaluate the efficiency and time cost of different approaches.

\textbf{S-Time} (Steady-state Time) denotes the average time required for all grasping processes to reach force steady states from the beginning. The force steady state refers to the fluctuation interval in which the value of the contact force maintains a distance from the end state that does not exceed ±5\% of the maximum range of variation. The steady-state time of a single grasp is the moment when more than 80\% of the forces enter the steady state. 

\begin{figure}[!t]
	\centering
	\includegraphics[width=3.2in]{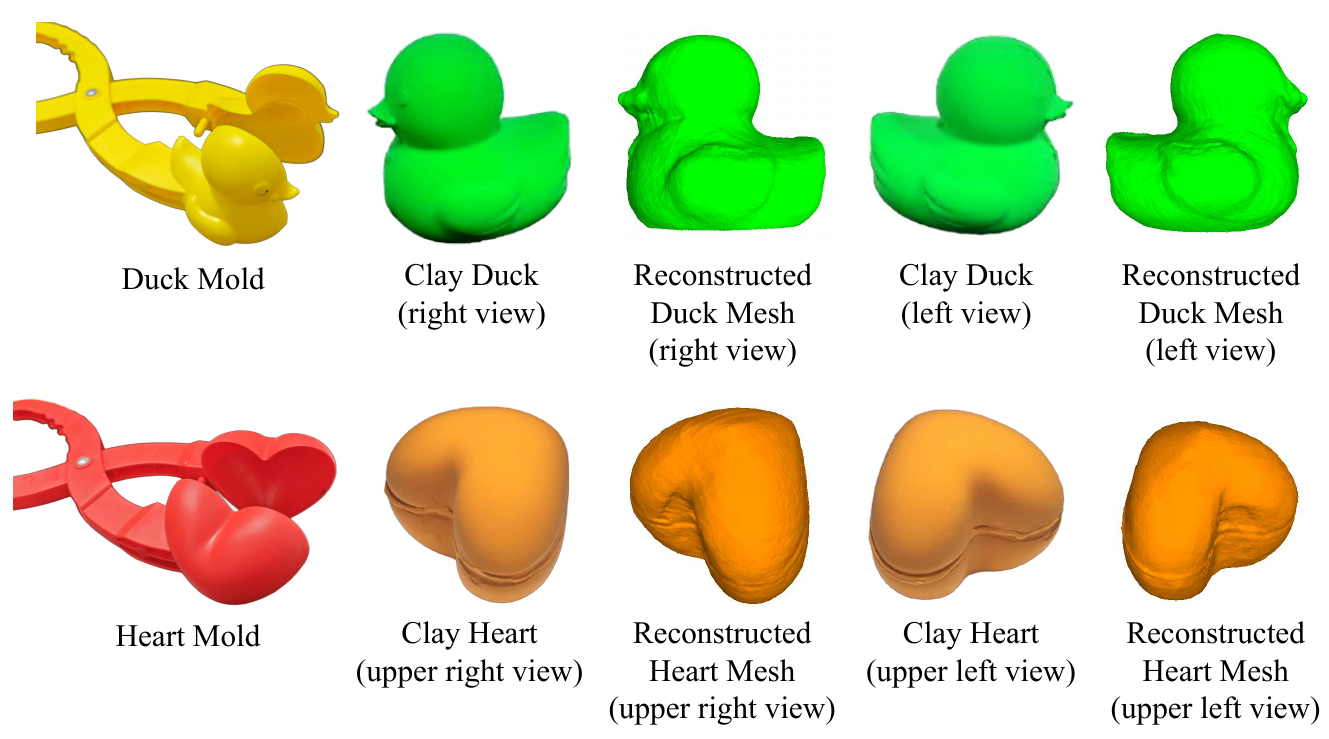}%
	\caption{Deformation assessment specimens and reconstructed meshes.}
	\label{fig_9: Reconstruction example.}
	\vspace{-10pt}
\end{figure}

\textbf{P-Time} (Processing Time) refers to the average time required from the input of perceptual features to the output of action commands.

\textbf{DCD} (Deformation Chamfer Distance). Besides direct force evaluation metrics (FEMs), we utilize Deformation Chamfer Distance (DCD) to quantitatively assess the differences between pre-grasp and post-grasp reconstructed meshes of clay specimens. This approach takes advantage of the plastic deformation properties of clay, allowing us to accurately capture the maximum deformations caused by the grasping process. By using DCD, we can easily visualize and quantify the impact of excessive forces on the material. We perform reconstructions of the clay specimens before and after grasping by Instant-NGP~\cite{muller2022instant} to record the deformation. Specifically, images of objects are captured uniformly across the hemisphere using the camera mounted on the robotic arm. To achieve high-quality reconstruction results, we employ SAM-Track~\cite{cheng2023segment} to accurately segment the objects across multiple images and effectively eliminate irrelevant backgrounds. Subsequently, images are processed through Colmap~\cite{fisher2021colmap} to access pose estimation. Finally, meshes of objects are generated within one minute through the training and rendering process. The Chamfer Distance between the normalized point clouds obtained pre-grasp and post-grasp is used to quantify the deformation. Given a successful grasp, a smaller deformation correlates with a more appropriate application of grasping forces. Fig.~\ref{fig_9: Reconstruction example.} illustrates clay duck and clay heart specimens manufactured using molds, along with their corresponding reconstructed meshes before being grasped. The DCD metric is formally calculated as the mean normalized Chamfer Distance between pre-grasp and post-grasp reconstructed meshes, averaged across three grasping trials for each specimen type. Consequently, each approach's DCD evaluation comprises six complete assessment cycles involving grasping, mesh reconstruction, and distance computation.

\begin{table*}[!t]  
	\caption{Evaluation of Different Approaches}  
	\label{tab_3:comparisons}  
	\centering
	\begin{threeparttable}
		\begin{tabular}{>{\centering\arraybackslash}m{2.8cm}>{\centering\arraybackslash}m{1.1cm}>{\centering\arraybackslash}m{2.2cm}>{\centering\arraybackslash}m{1.3cm}>{\centering\arraybackslash}m{1.6cm}>{\centering\arraybackslash}m{1.5cm}>{\centering\arraybackslash}m{1.3cm}>{\centering\arraybackslash}m{2cm}}  
			\hline  		
			\textbf{Approaches} & \textbf{SR}$\uparrow$ & \textbf{FEM-AT}$\downarrow$  & \textbf{FEM-KE}$\downarrow$ & \textbf{FEM-MAE}$\downarrow$ & \textbf{S-Time}$\downarrow$ & \textbf{P-Time}$\downarrow$ &\textbf{DCD in $10^{-3}$}$\downarrow$ \\
			\hline
			\textbf{Teleoperation} & 70.33\% & 536.88 ± 178.81 N & 1.53 N & 19.32 N & 6.59 s & — & 29.79 \\
			\textbf{Admittance Control} & 84\% & 824.36 ± 247.95 N & 3.38 N & 19.33 N & \textbf{3.08 s} & — & 52.27 \\
			\textbf{Modified ACT}~\cite{zhao2023learning} & 66.67\% & 532.77 ± 128.16 N & 2.38 N & 19.32 N & 6.78 s & 16.53 ms & 28.12 \\
			\textbf{Modified MULSA}~\cite{li2022seehearfeel} & 51\% & 528.73 ± 118.58 N & 1.93 N & 19.32 N & 6.16 s & 31.94 ms & 35.86 \\
			\textbf{GenDexGrasp} & 59\% & 637.28 ± 209.02 N & 2.33 N & 19.32 N & 3.34 s & 14.57 s & 37.61 \\
			\textbf{TK-STGN} (ours) & \textbf{91.67\%} & \textbf{437.47 ± 61.29 N} & \textbf{1.42 N} & \textbf{8.44 N} & 3.17 s & \textbf{12.39 ms} & \textbf{18.85} \\
			\hline  
			\textbf{Human Operation} & 100\% & 417.39 ± 66.46 N & 1.21 N & 6.89 N & 2.43 s & — & — \\
			\hline  
		\end{tabular} 		
		\begin{tablenotes}  
			\footnotesize  
			\item \textbf{Note:} \textbf{FEM} stands for force evaluation metric. The detail of FEMs can be found in Sec.~\ref{sec:force_eval_metrics}. The upward arrows indicate that higher values are preferable, while the downward arrows signify that lower values are desirable. The same notation also applies to the table below.  
		\end{tablenotes}
	\end{threeparttable}
	\vspace{-8pt}
\end{table*}  

\subsection{Baseline Approaches}\label{sec:Baseline Approaches}

To demonstrate the superiority of our method, we compare it with existing different types of grasping methods as follows:

\textbf{Human Operation}. This is performed directly by the demonstrator with his hands while possessing a priori knowledge of the object's properties.
It serves as the source of demonstration data and represents the upper performance for all imitation learning-based grasping approaches. Its inclusion in the table provides a benchmark for evaluating grasp success rate, contact force, and grasping efficiency. 
Beside Human Operation, all other tested approaches are carried out using robotic hands.

\textbf{Teleoperation.} This is achieved by the demonstrator under limited observation, prior knowledge, and force feedback. We establish a bilateral mapping channel between the Dexmo glove\footnote{\url{https://www.dextarobotics.com/}} and the Inspire robotic hand, which includes a morphological mapping from the human interface to the remote system and a tactile mapping from the remote system to the human interface. As a human-in-the-loop robotic hand control approach, teleoperation with force feedback channels presents a feasible and highly controllable solution for implementing grasping with robotic hands. This approach integrates human instruction and robotic proprioception, ensuring controllability and adaptability in robotic operations across varied environments. When dangerous environments prevent human on-site operation and fully autonomous control lacks reliability, teleoperation becomes an ideal alternative and has been widely used in various scenarios~\cite{li2022dexterous,li2021survey,mizrahi2024telefmg}.

\textbf{Admittance Control}. In this method, we set the desired torque to 80\% of the robotic hand's maximum torque to ensure safety and enhance the grasping success rate. Simultaneously, we output position control commands while limiting the maximum closure angle. Each finger is controlled independently in this approach, disregarding coordination between finger movements. 

\textbf{Modified ACT}~\cite{zhao2023learning}. It maintains the original backbone structure while substituting its input and output with kinesthetic and tactile representations aligned with TK-STGN. This adaptation enables the migration of the ACT framework to our grasping task while providing a benchmark to specifically evaluate the performance of non-graph-based networks and our graph-based TK-STGN architecture.

\textbf{Modified MULSA}~\cite{li2022seehearfeel}. For this method, we preserve the backbone structure and implement an end-to-end mapping from multimodal perceptions (i.e., visual, kinesthetic, and tactile) to direct motion commands for the robotic hand through imitation learning. The demonstration data for the modified MULSA need to be reacquired through teleoperation to obtain additional required visual demonstrations and corresponding motion of the robotic hand. This baseline specifically evaluates the impact of distinct sensory modalities in demonstration data on grasping performance.

\textbf{GenDexGrasp}~\cite{li2023gendexgrasp}. This approach generates contact maps based on the point cloud of the grasped object, then constructs and solves an optimization problem targeting these contact maps to obtain joint angles for robotic hands, thereby accomplishing grasping tasks. This baseline approach enables performance comparison between the vision-based optimization approach and the tactile-kinesthetic-based imitation learning approach in grasping tasks.

\subsection{Comparative Evaluation and Ablation Experiments}\label{sec:Comparative Evaluation and Ablation Experiments}

We conduct a comparative analysis of various grasping approaches, with the evaluation results presented in Tab.~\ref{tab_3:comparisons}. 
Among all tested methods, excluding Human Operation, TK-STGN achieve the highest success rate, the lowest grasping force, and the greatest efficiency. Furthermore, the values for these metrics of TK-STGN closely align with those of the Human Operation. While the Admittance Control achieves a high success rate and grasping efficiency, it performs poorly in managing contact force. In contrast, Modified MULSA falls short in terms of success rates and grasping efficiency, but it performs well in contact force management. The failure of Admittance Control is mainly due to two factors: independent finger control and the lack of regulation for bending speed. Independent control can result in uncoordinated finger movements and disrupt the balance of force between the fingers. Additionally, excessive grasping speed can make slippery objects more likely to slip away.

\begin{table*}[!t] 
	\caption{Results of Ablation Experiments}  
	\label{tab_4:ablations}  
	\centering 
	\begin{threeparttable}
		\begin{tabular}{>{\centering\arraybackslash}m{2.1cm}>{\centering\arraybackslash}m{1.2cm}>{\centering\arraybackslash}m{2.2cm}>{\centering\arraybackslash}m{1.3cm}>{\centering\arraybackslash}m{1.6cm}>{\centering\arraybackslash}m{1.5cm}>{\centering\arraybackslash}m{1.3cm}>{\centering\arraybackslash}m{2cm}}  
			\hline  
			\textbf{Approaches} & \textbf{SR}$\uparrow$ & \textbf{FEM-AT}$\downarrow$  & \textbf{FEM-KE}$\downarrow$ & \textbf{FEM-MAE}$\downarrow$ & \textbf{S-Time}$\downarrow$ & \textbf{P-Time}$\downarrow$ &\textbf{DCD in $10^{-3}$}$\downarrow$ \\
			\hline  
			\textbf{T-GCN} & 74.33\% & 480.46 ± 119.98 N & 1.92 N & 19.32 N & 3.28 s & \textbf{1.32 ms} & 30.88 \\ 
			\textbf{K-GCN} & 77\% & 544.06 ± 156.99 N & 2.29 N & 19.32 N & 4.06 s & 1.95 ms & 36.07 \\
			\textbf{TK6-GCN} & 22\% & 472.13 ± 93.59 N & 1.55 N & 19.33 N & 6.42 s & 2.69 ms & 32.64 \\
			\textbf{TA-GCN} & 53\% & 431.44 ± 94 N & 1.58 N & 19.32 N & 3.83 s & 2.94 ms & 27.46 \\	
			\textbf{TK-GCN} & 86.33\% & 445.45 ± 109.56 N & 1.52 N & 8.54 N & 3.31 s & 2.83 ms & 24.91 \\	
			\textbf{TK-STGN} (ours) & \textbf{91.67\%} & \textbf{437.47 ± 61.29 N} & \textbf{1.42 N} & \textbf{8.44 N} & \textbf{3.17} s & 12.39 ms & \textbf{18.85} \\
			\hline  
		\end{tabular} 
	\end{threeparttable} 
	\vspace{-5pt}
\end{table*}

\begin{figure*}[!t]
	\centering
	\includegraphics[width=7.0in]{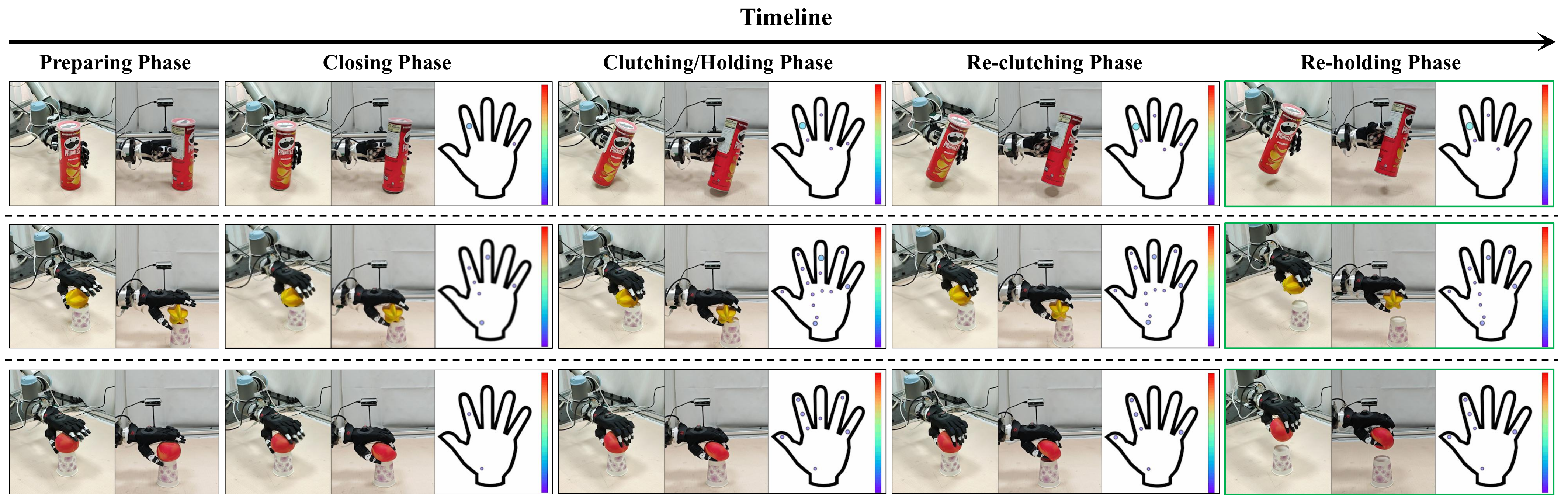}%
	\caption{The successful grasping processes using our TK-STGN. We present the grasping processes for three objects from two perspectives. Besides, we visualize the distribution of contact forces during the grasping process. The size and color of each point represent the magnitude of contact forces at corresponding locations, with values ranging from 0 N to approximately 20 N.}
	\label{fig_12:The successful grasping processes under the guidance of TK-GCN.}
	\vspace{-8pt}
\end{figure*}

The robotic hand's lower compliance and dexterity compared to the human hand often result in non-normal contact directions and misalignments between actual contact points and measurement locations. These factors introduce errors in tactile feedback when the glove is mounted on the robotic hand. Additionally, tangential stretching of the flexible substrate and bending of the tactile pads during the grasping process also introduce errors in measuring normal contact forces. In light of these challenges, comparisons between Modified ACT and TK-STGN demonstrate that our graph-based architecture is more robust to such tactile feedback deviations. TK-STGN notably achieves a higher grasp success rate and superior contact force management, utilizing the same input/output format and demonstration data. 
Compared to the Modified MULSA method, which requires processing additional visual features, our approach significantly reduces the processing time, enabling higher-frequency proprioceptive motion control. The success rates and force management results indicate that, upon reaching the predetermined position, the robotic hand can rely solely on tactile and kinesthetic feedback to achieve a stable grasp, and visual input is not a necessary reference.
As a pre-planning approach, GenDexGrasp directly generates target joint angles from point clouds through the contact map construction and force closure optimization phase. This process results in an average P-Time of 14.57 seconds, preventing real-time updates of the grasp strategy. In addition, its lack of real-time tactile feedback leads to significantly lower success rates and inferior contact force management compared to TK-STGN, strongly demonstrating the critical importance of tactile sensing for dynamic grasping tasks.

The comparison between Teleoperation and Human operation highlights the significant impact of limited observation perspectives, communication delays, and force feedback distortion on completing grasping tasks. This demonstrates that teaching via natural human operation is more efficient than teleoperation-based teaching and raises the upper limit of grasping performance. At least within our experimental setup, the TK-STGN approach learning from natural human operation outperforms the Teleoperation method in terms of grasping success rates and force management. These results suggest that our approach has the potential to replace human operators in executing tasks autonomously within complex and hazardous scenarios.

\begin{figure*}[!t]
	\centering
	\includegraphics[width=7.1in]{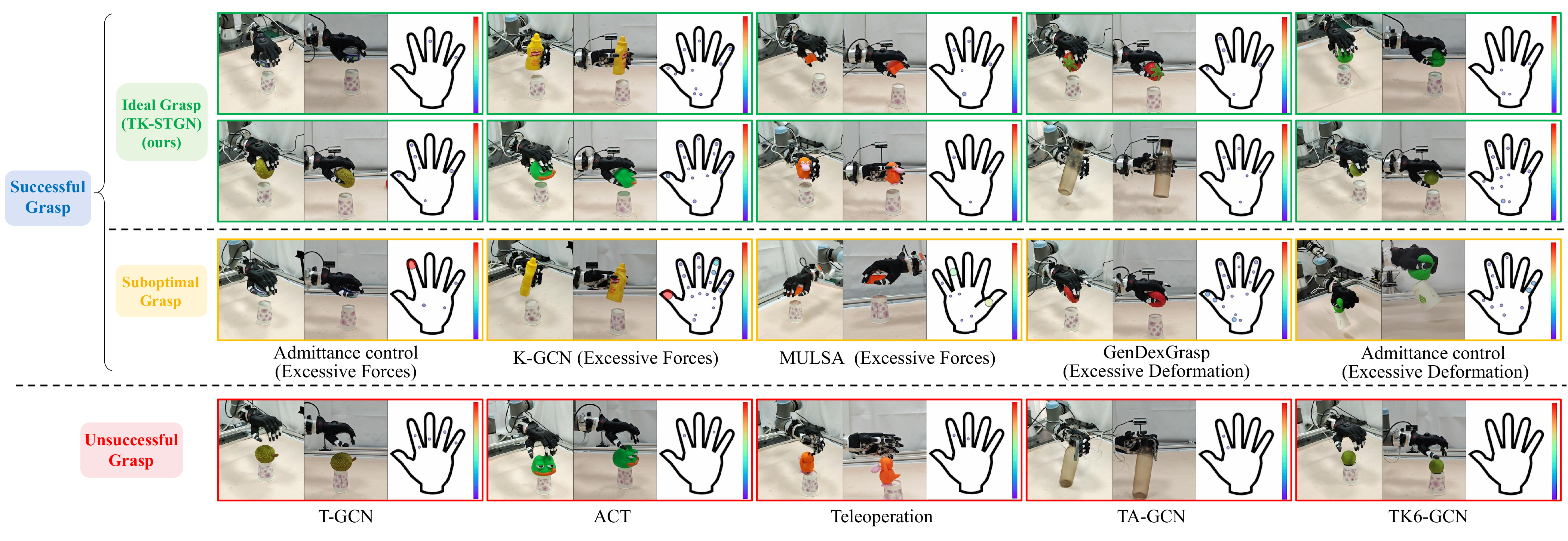}%
	\caption{Final states of ideal, suboptimal and unsuccessful grasps using different grasping approaches. The annotations below the subfigures identify the grasping approach and the reasons why this trial is categorized as successful yet suboptimal, including excessive force and excessive deformation. Note that the training and testing of the Modified MULSA are conducted based on the right hand, as the visual input capture device is fixed to the left side of the robotic hand.}
	\label{fig_10:Suboptimal and failed grasps under different grasping strategies.}
\end{figure*}

\begin{figure*}[!t]
	\centering
	\includegraphics[width=7.1in]{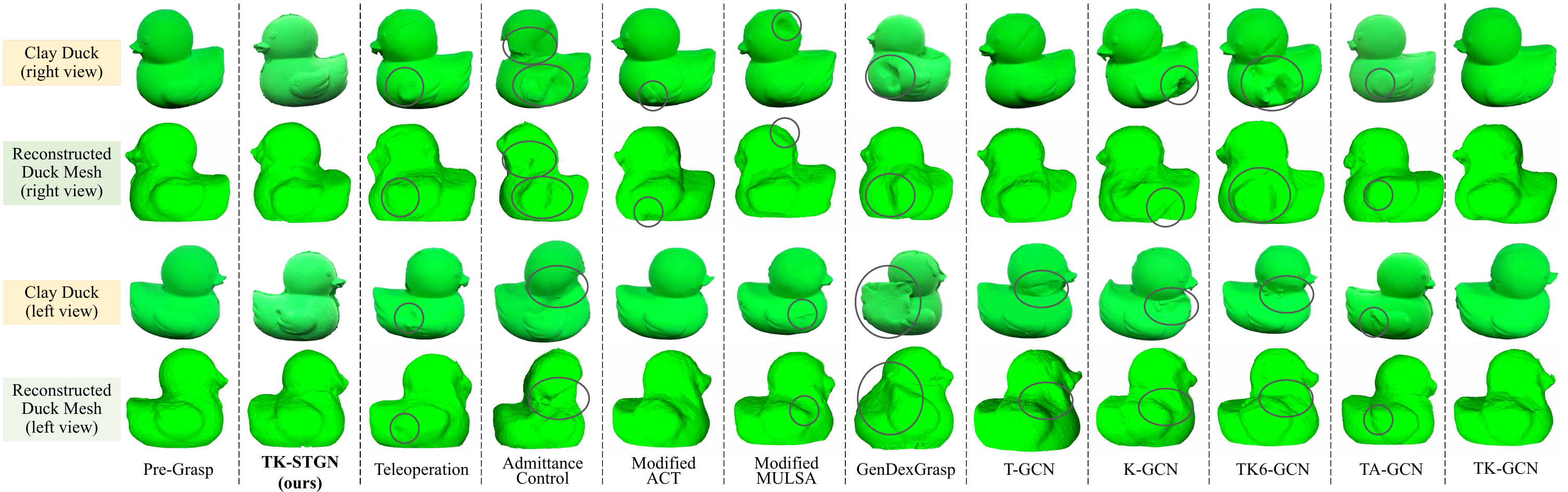}%
	\caption{Representative mesh reconstruction comparisons before and after grasping across different approaches (one specimen shown per approach from six trials each). Circles highlight major deformations.}
	\label{fig_11:Reconstruction results before and after grasping.}
\end{figure*}

Additionally, we perform ablation experiments on the TK-STGN to assess the impact of its individual components. The evaluation results of the ablation experiments are presented in Tab.~\ref{tab_4:ablations}. T-GCN and K-GCN utilize tactile features and kinesthetic features, respectively. Compared to TK-STGN, these approaches exhibit better efficiency, but they suffer from significantly reduced SRs and degraded performance in contact force management. During grasping scenarios, the actions controlled by K-GCN tend to converge to a fixed angle regardless of the object being grasped. Under T-GCN, actions often result in insufficient contact, causing slippage. 
TK6-GCN does not employ our proposed polar coordinate-based representation of finger movements. Instead, it uses the position and orientation of each joint in the inertial frame as input. This motion representation accurately reconstructs the motions of each finger joint. However, this over-parameterized description makes it more challenging to learn general grasping skills from multiple demonstrations. Despite our efforts to maintain consistent grasp positions and postures across trials, the fine details reveal larger differences between demonstrations, particularly when they include both side grasps and top-down grasps. Additionally, this representation renders TK6-GCN highly sensitive to input perturbations, resulting in oscillatory movements. Consequently, it experiences a significant drop in grasping success rate and requires a longer time to reach a steady state. 
In the TA-GCN approach, the well-established axis-angle representation replaces our kinesthetic representation. Axis-angle representation  encodes joint motions using a direction vector and angular displacement and is widely used in hand reconstruction. Compared to TK6-GCN and TA-GCN, our motion representation neglects the fine characterization of non-critical movements in grasping but incorporates flexion/extension velocities and bone lengths. This trade-off facilitates efficient skill transfer across diverse human demonstrators and robotic hands, yielding higher grasp success rates and more appropriate contact force management, as substantiated in Tab.~\ref{tab_4:ablations}. Unlike TK-STGN, TK-GCN eliminates the LSTM layers and attention mechanisms, which are used for extracting temporal features. Although TK-STGN exhibits longer P-time, it demonstrably increases success rates by 5.34\% while achieving superior contact force management.

\begin{figure*}[!t]
	\centering
	\includegraphics[width=7.1in]{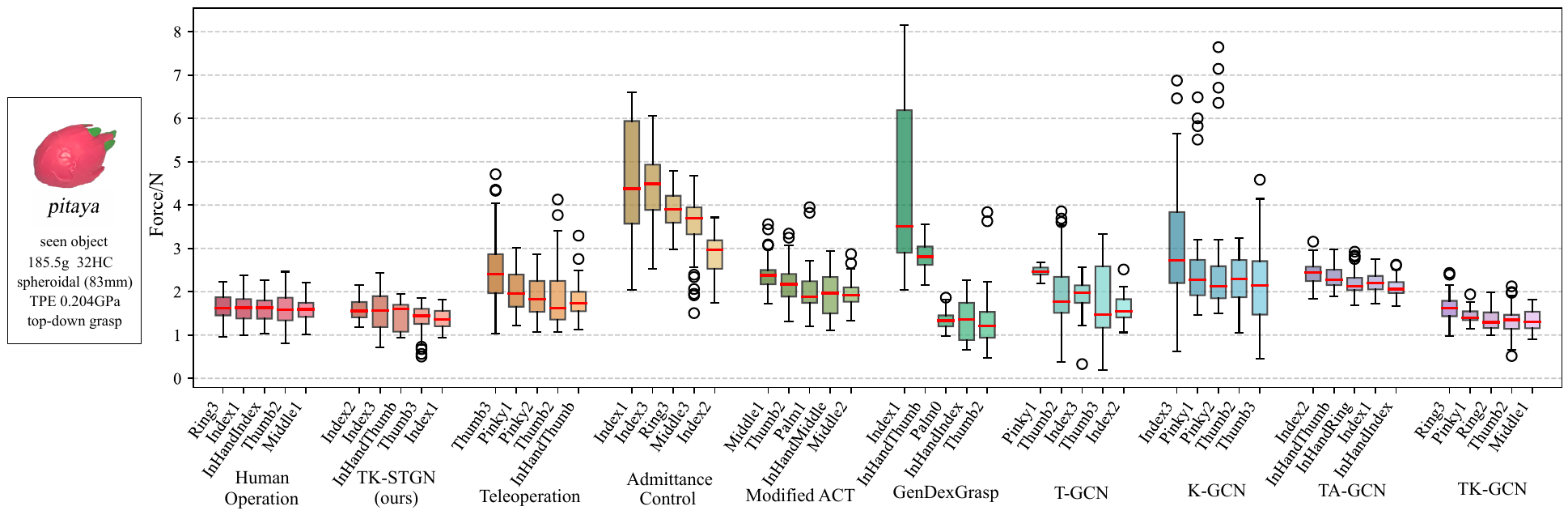}%
	\caption{Distribution of critical contact forces among 50 successful grasps for \textit{pitaya} at the terminal stage with various approaches. The top five grasping forces with the highest average amplitude among 50 successful grasps are identified as critical forces, and their corresponding labels are marked in the figure.}
	\label{fig_13a:Distribution of critical contact forces with various strategies}
\end{figure*}
\begin{figure*}[!t]
	\centering
	\includegraphics[width=7.1in]{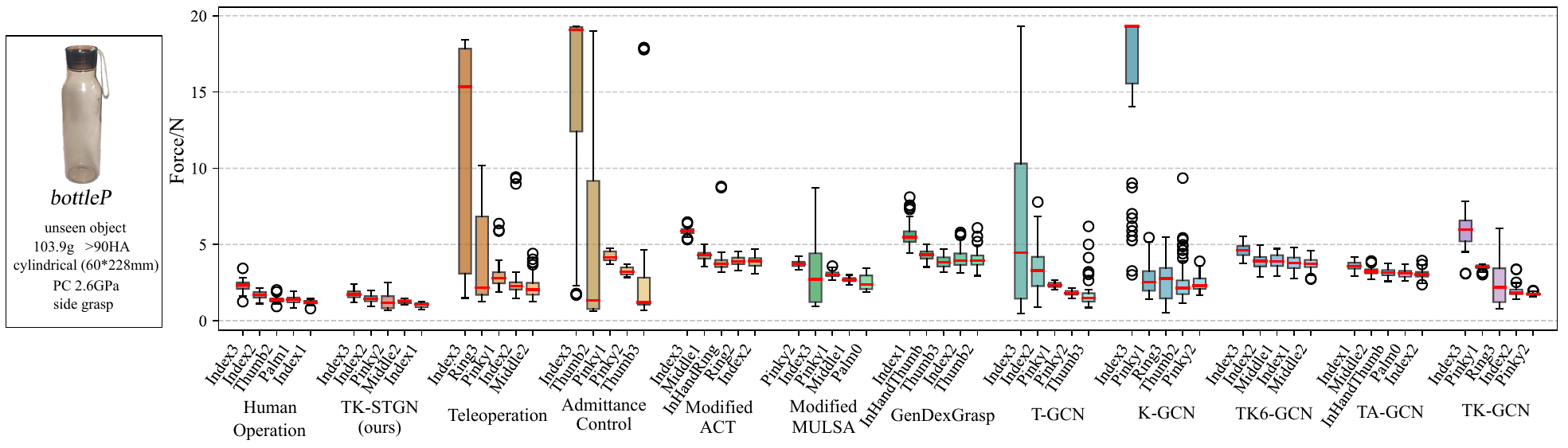}%
	\caption{Distribution of critical contact forces among 50 successful grasps for \textit{bottleP} at the terminal stage with various approaches.}
	\label{fig_13c:Distribution of critical contact forces with various strategies}
	\vspace{-8pt}
\end{figure*}

Following the analysis of the quantitative results, we present a visual depiction of the grasping performance.
Fig.~\ref{fig_12:The successful grasping processes under the guidance of TK-GCN.} shows some of the successful grasping processes using TK-STGN approach.
Fig.~\ref{fig_10:Suboptimal and failed grasps under different grasping strategies.} illustrates the final states of successful and unsuccessful grasps using different approaches.
We further classify successful grasp into ideal and suboptimal grasp in Fig.~\ref{fig_10:Suboptimal and failed grasps under different grasping strategies.}. Most suboptimal grasps succeed in preventing dropping but exhibit excessive forces or deformation due to poor force management. As shown in Tab.~\ref{tab_3:comparisons} and Tab.~\ref{tab_4:ablations}, while counted as successes, suboptimal grasps contribute to higher force evaluation values and are more likely to cause damage to fragile objects. Only grasps achieving both object retention and proper force management are defined as ideal grasps.
Fig.~\ref{fig_11:Reconstruction results before and after grasping.} depicts representative reconstruction outcomes of deformable clay ducks after being grasped by different approaches. The post-grasp clay duck visualizes the impact of poor contact force management. Through reconstructed meshes, we quantify this impact using DCD, as listed in Tab.~\ref{tab_3:comparisons} and Tab~\ref{tab_4:ablations}. Both visual and quantitative results demonstrate TK-STGN's outstanding performance in contact force management.

Fig.~\ref{fig_12:The successful grasping processes under the guidance of TK-GCN.}-\ref{fig_11:Reconstruction results before and after grasping.} highlight the capability of our TK-STGN approach in managing contact forces throughout the grasping process and preventing slippage on objects with diverse attributes. In contrast, other methods often result in uncoordinated finger movements, insufficient contact, excessive deformation or forces, and heightened sensitivity to disturbances.
Dynamic demonstrations of the grasping processes and further details are available in the supplementary video and via \url{https://grasplikehuman.github.io/.}

In addition, \textit{pitaya} and \textit{bottleP} are selected to assess the magnitude and consistency of critical contact forces in the terminal state using different grasping approaches. Fig.~\ref{fig_13a:Distribution of critical contact forces with various strategies} and Fig.~\ref{fig_13c:Distribution of critical contact forces with various strategies} show the distribution of these forces for grasping \textit{pitaya} and \textit{bottleP}, respectively. For each approach, data from 50 successful grasps are collected. However, due to the negligible success rate of using Modified MULSA and TK6-GCN for grasping the \textit{pitaya}, achieving 50 successful grasps for these objects is nearly impossible. Consequently, the corresponding data are absent in Fig.~\ref{fig_13a:Distribution of critical contact forces with various strategies}. Similar to the distribution of grasping contact forces observed in Human Operation, we posit that for a given object, the grasping forces should be minimized and remain consistent. The boxplot reflects this characteristic through a compact, lower-positioned interquartile range and fewer, more closely clustered outliers.

Among all successful grasps, the distribution of critical forces is relatively more concentrated and of lower magnitude in those controlled by our TK-STGN. In contrast, the distributions of critical forces under Teleoperation, Admittance Control, GenDexGrasp, K-GCN, and TK6-GCN are notably more dispersed and exhibit higher magnitudes. Although the Teleoperation process incorporates force feedback, it suffers from latency and inaccuracies as perceived by the operator, leading to inconsistent force distribution. Actions guided by the Admittance Control approach lack proper finger coordination, as each finger operates independently based on its desired contact force. This results in higher overall contact forces and greater variability in terminal states. GenDexGrasp and K-GCN approaches lack tactile perception entirely, resulting in poor force management capabilities. TK6-GCN's sensitivity to the inputs induces oscillations during grasping, leading to a more diverse and inconsistent force distribution.

In conclusion, for diverse objects and different grasping modes, our TK-STGN approach demonstrates the ability to maintain the highest grasp success rate while achieving elegant contact force management and coordinated finger movements throughout the grasping process. Our approach effectively avoids slipping, excessive deformation, oscillations, and excessive contact forces while delivering more consistent grasps for the same object. Compared to the other grasping approaches, our TK-STGN approach exhibits performance that more closely approximates human operation.

\subsection{Robustness and Generalization Evaluation}\label{sec:Robustness and Generalization Evaluation}

\begin{figure}[!t]
	\centering
	\includegraphics[width=3.4in]{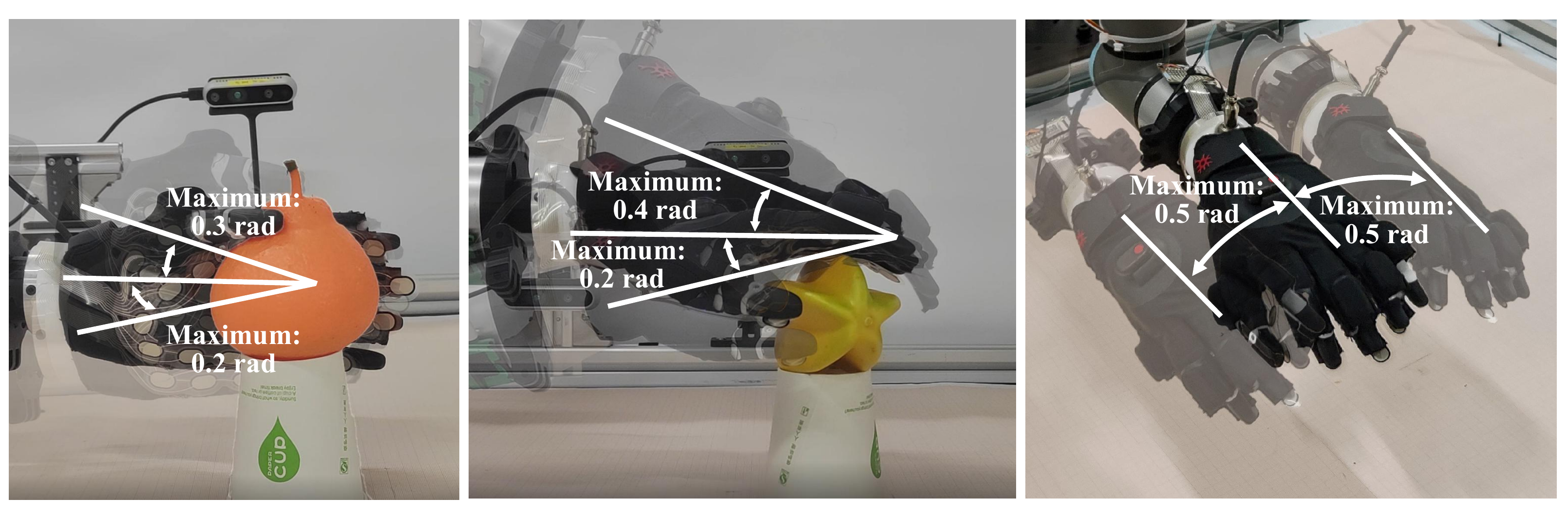}%
	\caption{The range of random approach angles. For side grasps, the maximum forward and backward pitch angles are 0.3 radians and 0.2 radians, respectively. For top-down grasps, the maximum forward and backward pitch angles are 0.4 radians and 0.2 radians, while the maximum roll angle is 0.5 radians.}
	\label{fig_16:The range of random approach angles.}
\end{figure}

\begin{figure}[!t]
	\centering
	\includegraphics[width=3.4in]{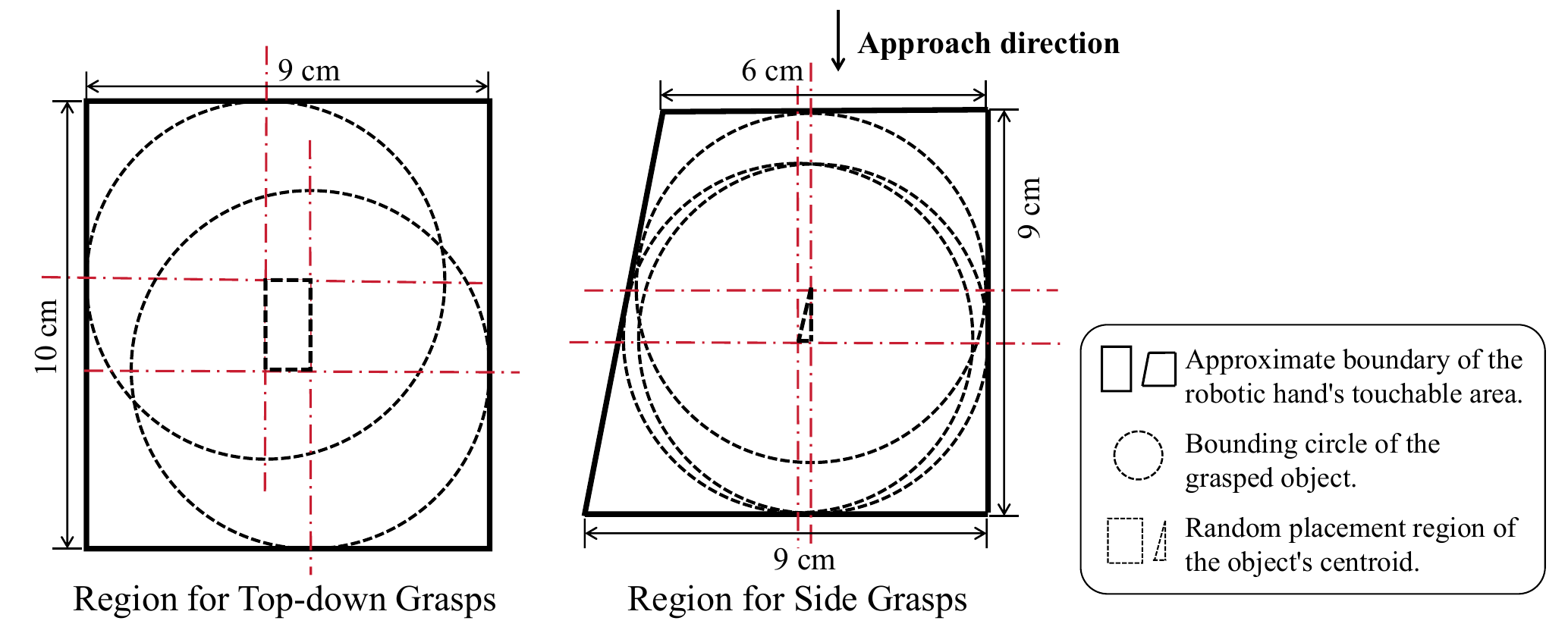}%
	\caption{Region for random placement of the grasped object (top-down view). The random selection region is determined by the grasping mode and the size of the robotic hand and manipulated objects. The approximate size of the touchable boundary marked in the figure is based on on-site measurements of the Inspire robotic hand.}	
	\label{fig_15:Region for random placement of the grasped object.}
	\vspace{-8pt}
\end{figure}

\begin{figure*}[!t]
	\centering
	\includegraphics[width=7.0in]{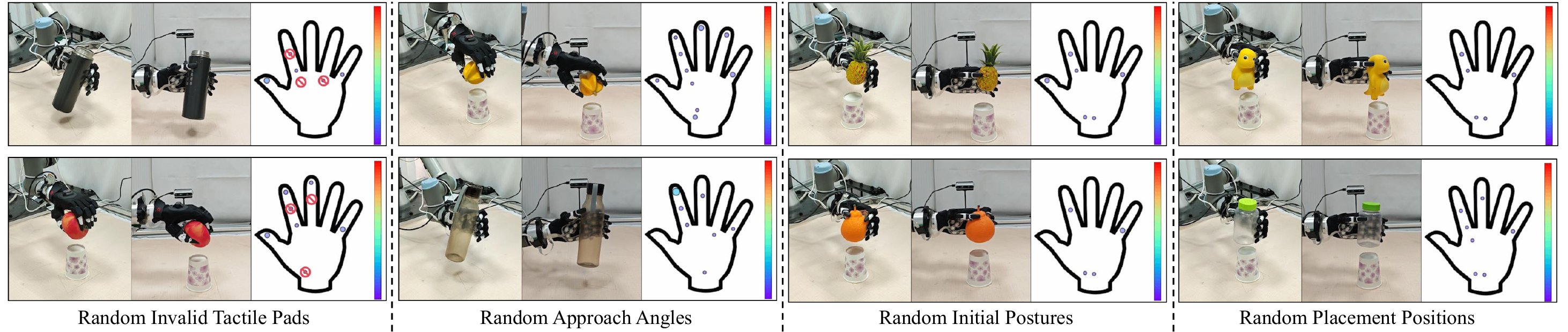}%
	\caption{Successful grasps under random setups. The labeled locations represent tactile sampling pads that are randomly masked.}
	\label{fig_17:Successful grasps under random conditions.}
\end{figure*}

In this section, we further examine the performance of TK-STGN approach, focusing on its robustness to random invalid tactile pads, random approach angles, random initial postures, and random placement positions. We also evaluate its generalization capability across different robotic hands. To maintain representativeness while ensuring experimental efficiency, all robustness and generalization tests are exclusively conducted using five seen objects and five unseen objects. As detailed in Sec.~\ref{sec:Comparative Evaluation and Ablation Experiments}, Modified ACT achieves the second-best overall performance among all imitation learning-based approaches and is surpassed only by our approach. Therefore, we select Modified ACT as a baseline for grasping tests under the same experimental conditions, providing a reference for assessing the robustness and generalization capabilities of our TK-STGN approach.

While testing with random invalid tactile pads, three channels are randomly turned off in each grasping trial to simulate data loss caused by tactile sensor damage or contact position shifts. As shown in Fig.~\ref{fig_16:The range of random approach angles.}, the range of random approach angles includes variability in pitch angles for side grasps and both pitch and roll angles for top-down grasps. In the random initial posture setting, each finger of the robotic hand is initialized to a random angle between 70 and 85 degrees before grasping. As for random placement positions, the initial position of the robotic hand remains consistent, while the target object is randomly positioned within a defined area, as illustrated in Fig.~\ref{fig_15:Region for random placement of the grasped object.}. The random placement areas for side and top-down grasps are specifically configured to ensure that the manipulated objects remain within the robotic hand's operational range. The random placement positions and approach angles simulate the effects of perception and execution errors on the initial grasp state when the robotic arm transitions to the preset grasping position.

The SRs in different randomized setups are shown in Tab.~\ref{tab_4:Success Rates Under Randomized Settings}. The SRs are calculated based on 15 grasps per object, totaling 150 grasps per setup. Compared to the baseline setup, the random invalid pads have little to no impact on the SR of our approach. However, for the Modified ACT approach, which shares the same input-output structure and demonstration data as our method, the SR drops significantly under the random invalid tactile pads setup. This highlights that the Modified ACT method is more sensitive to disrupted sensory information, whereas our approach maintains robust performance in such scenarios, thanks to TK-STGN's ability to handle individual anomalous inputs. Additionally, in other randomized experimental settings, the SRs reduction for the Modified ACT method is also significantly greater compared to our approach.

\begin{table}[!t]  
	\centering  
	\caption{Success Rates Under Randomized Settings}
	\begin{threeparttable}   
		\begin{tabular}{>{\centering\arraybackslash}m{3.5cm}>{\centering\arraybackslash}m{2cm}>{\centering\arraybackslash}m{2cm}}  
			\hline  
			\textbf{Robustness Testing Scenarios} & \textbf{Ours$\uparrow$} & \textbf{Modified ACT~\cite{zhao2023learning}$\uparrow$}\\ \hline
			Baseline &  93.33\% &  70\%\\
			Random Invalid Tactile Pads & 92.67\%\big|-0.67\% &  57.33\%\big|-12.67\% \\
			Random Approach Angles & 92\%\big|-1.33\% &  63.33\%\big|-6.67\% \\ 
			Random Initial Postures & 91.33\%\big|-2\% &  66.67\%\big|-3.33\%\\   
			Random Placement Positions & 86.67\%\big|-6.67\% &  54.67\%\big|-15.33\% \\ 
			
			\hline  
		\end{tabular}
		\begin{tablenotes}  
			\footnotesize  
			\item \textbf{Note:} The values after the vertical bars indicate the reduction in success rate compared to the baseline configuration for each random setup.
		\end{tablenotes}
	\end{threeparttable}
	\label{tab_4:Success Rates Under Randomized Settings}
	\vspace{-8pt}
\end{table}

In the case of random approach angles, our approach maintains a high level of grasp success because our proposed kinesthetic feature representation is unaffected by end-effector orientation perturbations. Random initial postures also affect the success rate. Still, the impact is limited since our demonstration process does not require identical initial postures, which indicates that our demonstration data already includes the corresponding action strategies. In comparison, the variability in placement positions exerts the most significant impact on the grasp success rate. The decline in SR is attributed to the difficulty in establishing force equilibrium due to the variability in placement positions. Such variability can cause the object to deviate from the robotic hand's grasping center or position it near the boundaries of the reachable workspace, thereby complicating the achievement of a stable force balance. Despite these challenges, TK-STGN approach manages to achieve a grasp success rate of 86.67\%. Fig.~\ref{fig_17:Successful grasps under random conditions.} illustrates the final states of the selected successful grasps under random conditions, including the distribution of contact forces at the corresponding moments.

\begin{figure*}[!t]
	\centering
	\includegraphics[width=7.1in]{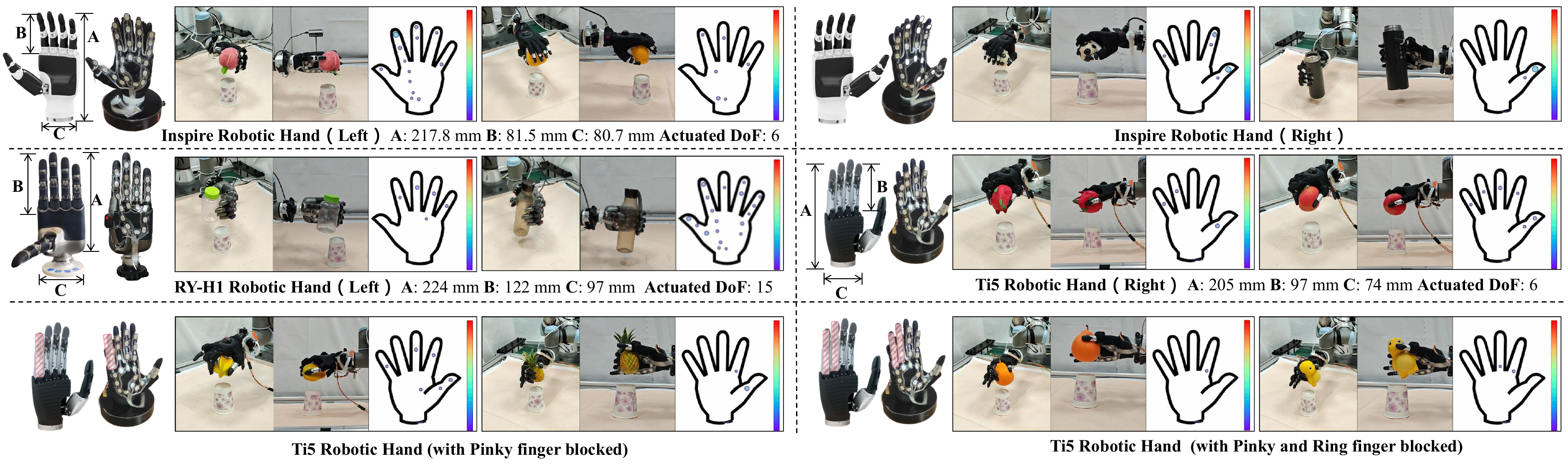}%
	\caption{Successful grasps with different robotic hands. Key dimensions and the number of actuated DoF of each hand are labeled.}
	\label{fig_18:Successful grasps with different hands.}
	\vspace{-8pt}
\end{figure*}

In addition, we test the generalization ability of our approach on different robotic hands, and the success rates are listed in Tab.~\ref{tab_5:Success Rate Using Different Manipulators.}. We also test the success rate when blocking the proprioception and motion of individual fingers.
Comprehensively, our cross-platform generalization evaluation encompasses six distinct robotic hand configurations. This evaluation involves variations across robotic platforms, the number of actuated DoF, the number of fingers, and left/right-handed variants of identical models. 
When employing different robotic hands, we only need to fine-tune the relevant parameters of the force-position hybrid mapping to ensure that the output dimensions and data format align with the control commands of the robotic hand. 

\begin{table}[!t]  
	\centering  
	\caption{Success Rates Using Different Robotic Hands} 
	\begin{threeparttable} 
		\begin{tabular}{>{\centering\arraybackslash}m{4cm}>{\centering\arraybackslash}m{0.8cm}>{\centering\arraybackslash}m{2.5cm}}  
			\hline  
			\textbf{Robotic Hands} & \textbf{Ours}$\uparrow$ & \textbf{Modified ACT}~\cite{zhao2023learning}$\uparrow$ \\ \hline
			Inspire Hand (Left) &  93.33\% &  70\%\\  
			Inspire Hand (Right) & 92.67\% &  71.33\%\\
			RY-H1 Hand (Left) & 92.67\% &  72.67\%\\
			Ti5 Hand (Right)& 94\%&  74\%\\
			Ti5 Hand (Right, with Pinky finger blocked) & 89.33\%&  66\%\\ 
			Ti5 Hand (Right, with Pinky and Ring finger blocked) & 84.67\% &  51.33\%\\ \hline  
		\end{tabular}  
	\end{threeparttable} 
	\label{tab_5:Success Rate Using Different Manipulators.}
	\vspace{-8pt}
\end{table} 

Fig.~\ref{fig_18:Successful grasps with different hands.} illustrates successful grasps achieved with various robotic hands, with the key dimensions and the number of acutated DoF labeled in the figure. The comparison of the SRs demonstrates that our approach exhibits strong generalization capabilities across different robotic hands. Our approach maintain relatively high grasp SRs even when proprioception and motion of 1-2 fingers are blocked. In contrast, the performance of the Modified ACT method shows a more significant decline under these conditions, highlighting its sensitivity to robotic hand configuration changes and inhibited finger functionality. Notably, the RY-H1 Hand\footnote{\url{http://ruirobot.cn/index.asp}} and Ti5 Hand\footnote{\url{https://www.ti5robot.com/en/}} achieve higher grasp success rates than the Inspire Hand in some configurations due to their larger finger length proportion, which facilitates the envelopment of grasped objects.

In summary, our approach demonstrates strong robustness to randomized experimental settings and superior generalization capability across different robotic hand configurations, significantly outperforming the Modified ACT method. The robustness tests indicate that our approach can achieve reliable grasping performance even with a certain level of perception and execution errors. Meanwhile, the generalization tests show that for grasping tasks, our approach can adapt to changes in robotic hand configurations without the need for re-collecting demonstration data or retraining the model.

\begin{table*}[!t]
	\caption{Changes in Success Rate with Expanding Training Datasets}  
	\label{tab_2:eva with expanding datasets}  
	\centering
	\begin{threeparttable}
		\begin{tabularx}{\textwidth}{>{\centering\arraybackslash}m{2.5cm}<{\centering}m{0.6cm}<{\centering}m{0.65cm}<{\centering}m{0.5cm}<{\centering}m{0.7cm}<{\centering}m{0.7cm}<{\centering}m{0.7cm}<{\centering}m{0.5cm}<{\centering}m{0.55cm}<{\centering}m{0.7cm}<{\centering}m{0.6cm}<{\centering}m{1.1cm}<{\centering}m{1.45cm}<{\centering}m{0.77cm}<{\centering}}
			\hline
			\multicolumn{1}{c}{\multirow{2}{*}{\parbox{2.5cm}{\centering \textbf{Objects in the Training Set}}}} & \multicolumn{10}{c}{\textbf{Successful Attempts/Total Attempts}}                                                                 & \multicolumn{1}{c}{\multirow{2}{*}{\parbox{1.1cm}{\centering \textbf{SR(Seen Objects)$\uparrow$}}}} & \multicolumn{1}{c}{\multirow{2}{*}{\parbox{1.45cm}{\centering\textbf{SR(Unseen Objects)$\uparrow$}}}} & \multicolumn{1}{c}{\multirow{2}{*}{\textbf{SR}$\uparrow$}} \\ \cline{2-11}
			\multicolumn{1}{c}{} & \multicolumn{1}{c}{\parbox{0.6cm}{\centering \textit{bottleS}}} & \multicolumn{1}{c}{\parbox{0.65cm}{\centering \textit{nailong}}} & \multicolumn{1}{c}{\parbox{0.5cm}{\centering \textit{pitaya}}}  & \multicolumn{1}{c}{\parbox{0.7cm}{\centering \textit{caram-bola}}} & \multicolumn{1}{c}{\parbox{0.7cm}{\centering \textit{football}}} & \multicolumn{1}{c}{\parbox{0.7cm}{\centering \textit{pine-apple}}} & \multicolumn{1}{c}{\parbox{0.5cm}{\centering\textit{jar}}} & \multicolumn{1}{c}{\parbox{0.55cm}{\centering\textit{citrus}}} & \multicolumn{1}{c}{\parbox{0.7cm}{\centering\textit{pome-granate}}} & \multicolumn{1}{c}{\parbox{0.6cm}{\centering\textit{bottleP}}} & \multicolumn{1}{c}{}                   & \multicolumn{1}{c}{}                   & \multicolumn{1}{c}{}                   \\ \hline
			
			[\textit{bottleS}]  & 13/15 & 3/15 & 1/15 & 0/15 & 3/15 & 4/15 & 3/15 & 8/15 & 2/15 & 14/15 & 86.67\% & 28.15\% & 34\% \\
			
			[\textit{bottleS},\textit{nailong}]  & 15/15 & 15/15 & 1/15 & 7/15 & 3/15 & 8/15 & 13/15 & 15/15 & 3/15 & 15/15 & 100\% & 54.17\% & 63.33\% \\
			
			[\textit{bottleS},\textit{nailong},\textit{pitaya}]  & 15/15 & 15/15 & 8/15 & 8/15 & 12/15 & 9/15 & 9/15 & 15/15 & 4/15 & 15/15 & 84.44\% & 68.57\% & 73.33\% \\
			
			[\textit{bottleS},\textit{nailong},\textit{pitaya}, \textit{carambola}(\textbf{L/R})] & 15/15 & 15/15 & 6/15 & 7/15 & 9/15 & 12/15 & 11/15 & 15/15 & 7/15 & 15/15 & 71.67\% & 76.67\% & 74.67\% \\
			
			[\textit{bottleS},\textit{nailong},\textit{pitaya}, \textit{carambola}(\textbf{L/R}), \textit{football}(\textbf{F/M})]& 14/15 & 15/15 & 9/15 & 12/15 & 14/15 & 13/15 & 15/15 & 14/15 & 11/15 & 15/15 & 85.33\% & 90.67\% & 88\% \\
			\hline  
		\end{tabularx}  
		\begin{tablenotes}  
			\footnotesize  
			\item \textbf{Note:} \textit{carambola}(\textbf{L/R}) indicates that the dataset of \textit{carambola} comes from the \textbf{left} and \textbf{right} hands of the same demonstrator;  \textit{football}(\textbf{F/M}) indicates that the dataset of \textit{football} comes from different \textbf{female} and \textbf{male} demonstrators.  
		\end{tablenotes}
	\end{threeparttable}
\end{table*}

\section{Discussion}

\subsection{Training with Expanding Datasets}\label{sec:Training with Expanding Datasets}

Here, we explore the relationship between the training set scale and its impact on grasp success and model generalization. Tab.~\ref{tab_2:eva with expanding datasets} lists the objects included in each training set, along with the number of successful attempts and success rates for the seen and unseen objects. Notably, the data for the \textit{carambola} is derived from the same demonstrator using both the left and right hands; the data for the \textit{football} is sourced from a male demonstrator with a hand length of 18.5 cm and a female demonstrator with a hand length of 15.5 cm.

As observed in Tab.~\ref{tab_2:eva with expanding datasets}, adding new objects to the training set improves grasp success rates not only for those objects but also for others with similar properties, such as size, grasp mode, and hardness. Notably, incorporating demonstration data from both left and right hands, as well as from diverse demonstrators, leads to a steady increase in success rates for unseen objects and overall performance. This highlights the compatibility of our data glove and kinesthetic-tactile representation with heterogeneous demonstrations. Moreover, the developed demonstration dataset for generalizable grasping demonstrates significant potential for large-scale expansion.

\begin{figure*}[!t]
	\centering
	\includegraphics[width=7.0in]{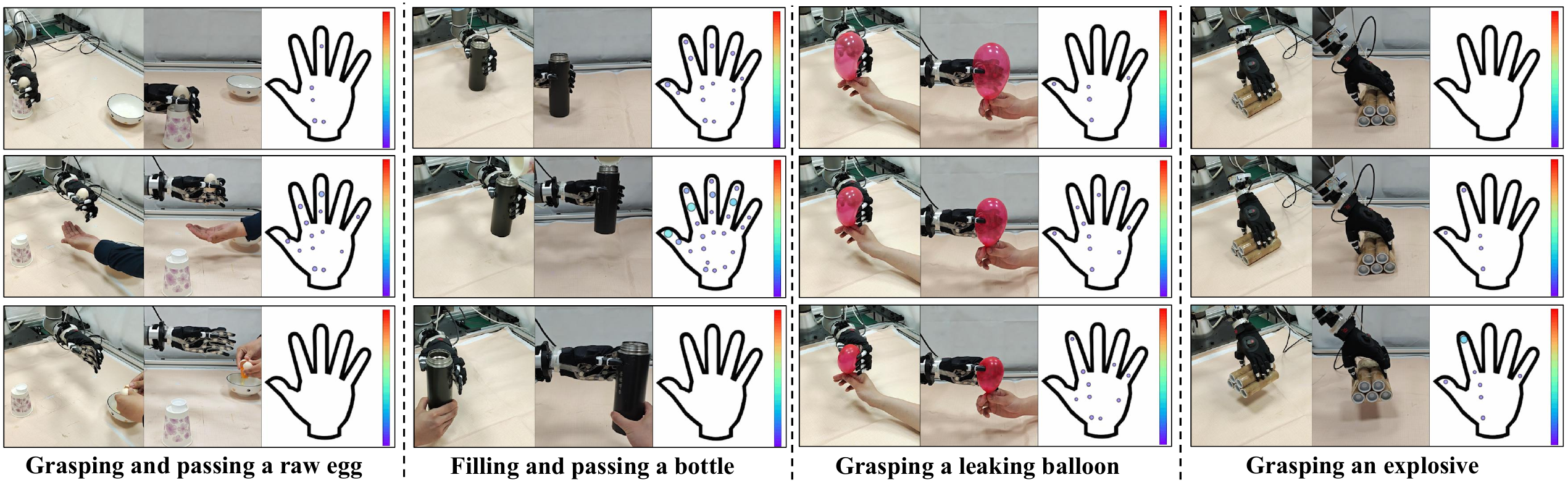}%
	\caption{Grasp-based task execution processes and the corresponding contact force distribution.}
	\label{fig_20:Specific grasping-based tasks.}
\end{figure*}

The fluctuations in success rates as the training set expands are primarily observed with objects such as \textit{pitaya}, \textit{carambola}, \textit{football}, and \textit{jar}. These objects present greater challenges for grasping, making their success rates particularly sensitive to the expanding dataset. Specific difficulties include insufficient friction during top-down grasps due to deformable characteristics (\textit{pitaya}), inconsistent tactile feedback caused by irregular shapes leading to random contact points (\textit{carambola}), limited contact area due to small size (\textit{football}), and a tendency to slip due to slippery surfaces (\textit{carambola} and \textit{jar}).
\vspace{-5pt}
\subsection{The Role of Tactile Feedback in Grasping}

Tactile perception plays a crucial role in human grasping, especially when visual feedback is unavailable. Our approach achieves successful grasping based solely on real-time proprioceptive sensory inputs without any prior knowledge or visual observation of the objects. It ensures effective contact force management and finger coordination, achieving performance that closely resembles human hand grasping.

As shown in Tab.~\ref{tab_4:ablations} and Fig.~\ref{fig_13a:Distribution of critical contact forces with various strategies}-\ref{fig_13c:Distribution of critical contact forces with various strategies}, the grasp success rates for K-GCN and T-GCN are comparable when relying solely on kinesthetic or tactile perception, respectively. However, T-GCN demonstrates superior contact force management, evidenced by smaller FEM values and a narrower, more concentrated contact force distribution during repeated grasps on specific objects. When both tactile and kinesthetic perceptions are integrated, the grasp success rate significantly increases, and contact force management is also improved. This enhancement is attributed to the combined information from contact force and joint angle variations, which provides a reliable estimation of the object's hardness or elastic modulus, thereby facilitating more effective adjustments to the grasp strategy. 
The comparison between GenDexGrap and our approach also highlights the key influence of real-time tactile feedback.
Additionally, the comparison in Tab.~\ref{tab_3:comparisons} between our approach and the Modified MULSA suggests that real-time visual feedback is optional once the robotic hand has been guided to a suitable pre-grasp position.

However, our approach utilizes a reduced-dimensional yet reasonable tactile representation, emphasizing the critical role of low-dimensional full-hand tactile feedback in grasping. Compared to our tactile representation, human tactile perception exhibits higher density and includes additional capabilities such as thermal detection and lateral shear strain perception. These capabilities are essential for implementing more complex dexterous manipulation with robotic hands.
\vspace{-5pt}
\subsection{The Generalizability and Robustness of Our Approach}

We demonstrate the generalization and robustness of our method in four aspects: compatibility with different demonstrators, various objects, random experimental setups, and different robotic hands. Due to our motion representation based on polar coordinates, which account for the skeletal models of different demonstrators' hands, our method can extract universal motion experience from the grasps of diverse demonstrators. Results in Tab.~\ref{tab_2:eva with expanding datasets} confirm our method's compatibility with different demonstrators. While training with five objects, our approach reliably grasps at least twenty objects with varying shapes, masses, and hardness. Additionally, we test our method by grasping clay specimens. Based on the DCD comparison shown in Tab.~\ref{tab_3:comparisons} and Fig. \ref{fig_11:Reconstruction results before and after grasping.}, we show that our approach can achieve reliable grasps with minimal deformation of the object's shape. 
Sec.~\ref{sec:Robustness and Generalization Evaluation} validates our approach's robustness to various random experimental conditions and its generalization capability across different robotic hands. Even when some fingers are disabled, i.e., when the five-fingered hand is reconstructed into four- or three-fingered hands, our approach adapts effectively while maintaining a high success rate.

We further evaluate our approach on several practical grasp-based tasks, including grasping and passing a raw egg, filling and passing a bottle, grasping a leaking balloon, and grasping an explosive. Fig.~\ref{fig_20:Specific grasping-based tasks.} illustrates the task execution process and the corresponding contact force distribution at different moments. Specifically, grasping and passing a raw egg tests the grasp force management for small, fragile objects under the disturbance caused by robotic arm movements. The water filling and passing task examines the grasp's robustness under changing object mass and arm-induced disturbances. Grasping a leaking balloon evaluates the compliance of the grasp and the ability to dynamicly manage contact forces. Grasping an explosive assesses the reliability of the grasp and force management under specialized grasp postures. As shown in Fig.~\ref{fig_20:Specific grasping-based tasks.}, our approach successfully completed all these tasks, demonstrating reliable grasps with well-coordinated finger movements and appropriate contact force management throughout the process. These results highlight the robustness and generalization capabilities of our approach in real-world application scenarios. More details are shown in the supplementary video. 
\vspace{-5pt}
\subsection{Limitations and Future Works}

As a preliminary exploration of glove-mediated skill transfer via proprioceptive feedback, multimodal graph representation, and natural human demonstrations, our approach exhibits limitations in four aspects: i) sparse tactile sensing, ii) simplified motion representation, iii) hardware-specific tuning challenges, and iv) constrained task design.

{i) Our current approach uses sparse tactile sampling points that only measure normal contact forces. Such sparse sensing does not fully capture the richness of human tactile sensation, which includes the ability to sense vibration, temperature, deformation, and lateral strain. However, we have verified that extracting the spatial topological correlation of sparse but palm-wide distributed tactile features also contributes to improved grasping performance. Specifically, the sparse tactile representation simplifies the learning problem, emphasizes the role of spatial topological features, and reduces the amount of demonstration data needed. Future improvements in mimicking human-like tactile feedback could be supported by commercial advancements in the development of denser, miniaturized, and modular tactile sensors. We plan to integrate these advanced high-density tactile sensors (e.g., BioTac or GelSight) with a hierarchical graph structure, where local tactile patterns are processed via convolutional operators or weighted edges to capture fine‐grained spatial relationships, before being incorporated into higher‐level vertices for skill transfer.

{ii) For grasping-centric tasks, we exclude the motion representation for finger abduction and adduction (except for the thumb), thereby accelerating training and enhancing robustness. However, a more detailed and precise motion representation could be beneficial for supporting complex manipulation, particularly for multi-DoF robotic hands advance.

{iii) We set TK-STGN's outputs as the desired joint angles and contact forces, which simplifies training and enables cross-platform transfer without data recollection and model retraining. However, current force-position hybrid mapping requires manual parameter tuning specific to the hardware. We plan to replace this with an optimization-based procedure for matching finger pose and contact force for adaptive mapping.

{iv) Our validation focused solely on grasping with pre-positioned hands. Future implementations may consider incorporating in-hand manipulation/adjustment tasks or integrating arm motion planning and multi-hand coordination.

In summary, our future work will focus on advancing glove-mediated skill transfer through high-density multimodal tactile sensing upgrades, fine-grained motion representation refinements that encompass full dexterous movements of human hands, and integrated arm-hand coordination for complex manipulation. Additionally, we will complete a comprehensive grasp evaluation using the YCB dataset~\cite{calli2015benchmarking} and develop a rigorous benchmark.

\section{Conclusion}

In summary, we develop a glove-mediated grasp skill transfer framework based on imitation learning that enables the transition from human proprioceptive operation to robotic hand manipulation. We integrate a data glove to collect kinesthetic and tactile features applicable to both human and robotic hands. Moreover, we propose a unified graph-based multimodal feature representation method. By leveraging human hand grasp demonstration data and the TK-STGN, our approach enables grasping objects with various properties, including deformable, slippery, and deformable ones. Comparative experiments illustrate the advantages of our approach in terms of grasp success rate, finger coordination, contact force management, as well as both grasp and computational efficiency. The results show that learning from natural human operations and the interaction between human hands and the real-world environment is both effective and efficient in achieving reliable grasping.
Additionally, ablation studies highlight the crucial role of tactile feedback in ensuring successful grasping and effective force management. Furthermore, our approach exhibits strong robustness to randomized experimental settings, indicating that it achieves reliable grasping performance even in the presence of certain levels of perception and execution errors. Meanwhile, generalization tests demonstrate that our approach can adapt to different robotic hand configurations for grasping tasks without the need to re-collect demonstration data or retrain the model.

\bibliographystyle{IEEEtran}

\bibliography{reference}

\vspace{-15pt}
\begin{IEEEbiography}[{\includegraphics[width=1in,height=1.25in,clip,keepaspectratio]{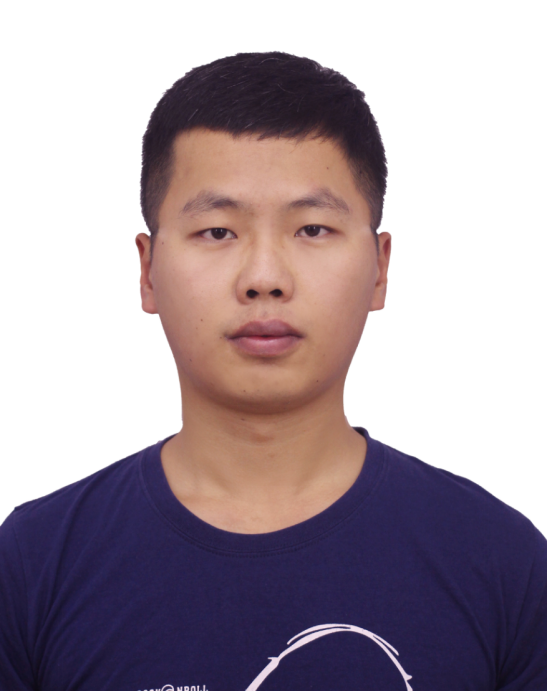}}]{Ce Guo}
	received the B.E. degree in automation and the M.E. degree in control science and engineering from the College of Intelligence Science and Technology, National University of Defense Technology. He is currently pursuing the Ph.D. degree. His research interests include imitation learning and robot manipulation.
\end{IEEEbiography}
\vspace{-25pt}
\begin{IEEEbiography}[{\includegraphics[width=1in,height=1.25in,clip,keepaspectratio]{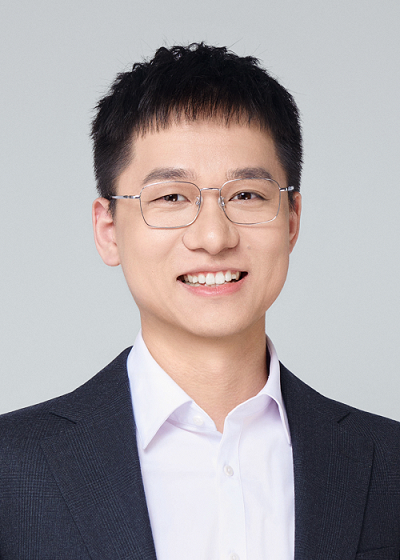}}]{Xieyuanli Chen} is now an Associate Professor at the National University of Defense Technology, China. He received his Ph.D. degree at Photogrammetry and Robotics Laboratory, the University of Bonn. He received his Master degree in Robotics in 2017 at the National University of Defense Technology, China. He received his Bachelor degree in Electrical Engineering and Automation in 2015 at Hunan University, China. He also serves as the Associate Editor for IEEE Robotics and Automation Letters.
\end{IEEEbiography}
\vspace{-25pt}
\begin{IEEEbiography}[{\includegraphics[width=1in,height=1.25in,clip,keepaspectratio]{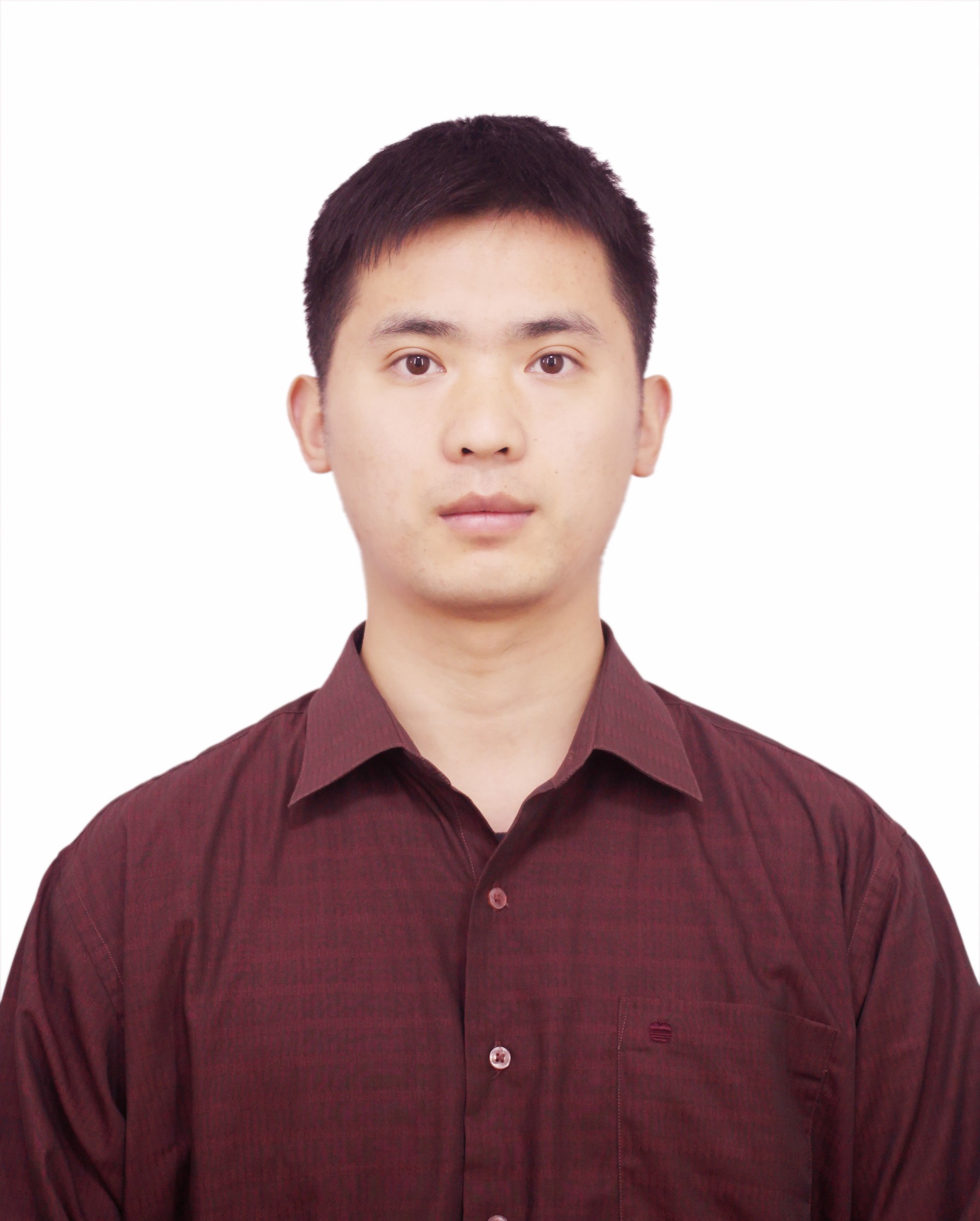}}]{Zhiwen Zeng}
	received his Ph.D. (2016) from the Robotics Research Center (RRC), College of Intelligence Science and Technology, National University of Defense Technology, and received his B.S. degree from the University of Electronic Science and Technology of China in 2009, and M.S. degree from National University of Defense Technology in 2011. From 2019, he serves as an associate professor on Robotics and Cybernetics. His research interests include path planning, human-robot interaction, multi-robot coordination distributed control, estimation of networked dynamical systems and the application to intelligent and robotic system.
\end{IEEEbiography}
\vspace{-25pt}
\begin{IEEEbiography}[{\includegraphics[width=1in,height=1.25in,clip,keepaspectratio]{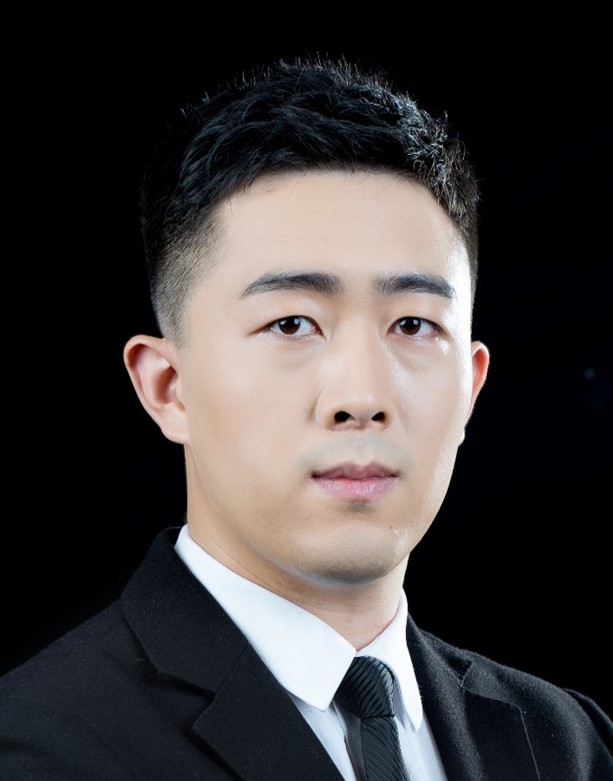}}]{Zirui Guo}
	received the B.E. degree in automation and the M.E. degree in control science and engineering from the College of Intelligence Science and Technology, National University of Defense Technology. He is currently pursuing the Ph.D. degree. His research interests include robot perception and manipulation.
\end{IEEEbiography}
\vspace{-25pt}
\begin{IEEEbiography}[{\includegraphics[width=1in,height=1.25in,clip,keepaspectratio]{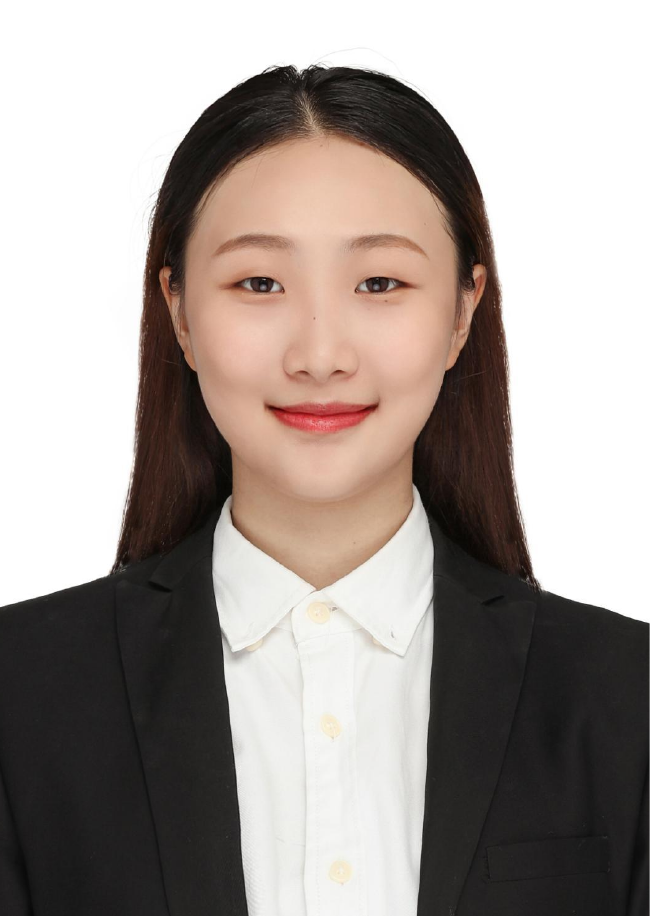}}]{Yihong Li}
	received a B.E. Degree in Vehicle Engineering from Fuzhou University, Fuzhou, China, in 2022. 
	She is now a graduate student at the National University of Defense Technology, investigating grasp control algorithms for robots with anthropomorphic robotic hands.
\end{IEEEbiography}
\vspace{-25pt}
\begin{IEEEbiography}[{\includegraphics[width=1in,height=1.25in,clip,keepaspectratio]{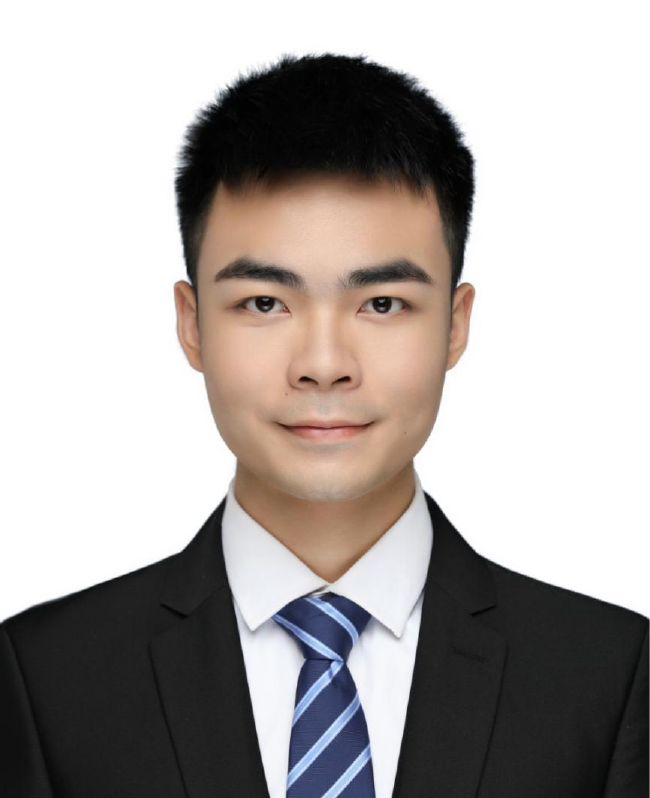}}]{Haoran Xiao}
	received the B.E. degree in intelligence science and technology from the College of Computer Science and Electronic Engineering, Hunan University. He is currently pursuing the Ph.D. degree. His research interests include robotic manipulation, and robotic motion and task planning.
\end{IEEEbiography}
\vspace{-25pt}
\begin{IEEEbiography}[{\includegraphics[width=1in,height=1.25in,clip,keepaspectratio]{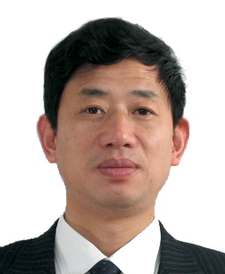}}]{Dewen Hu}
	received the B.S. and M.S. degrees in automatic control from Xi’an Jiaotong University, Xi’an, China, in 1983 and 1986, respectively, and the Ph.D. degree in automatic control from the National University of Defense Technology, Changsha, China, in 1999. He is currently a Professor with the College of Intelligence Science and Technology, National University of Defense Technology. From 1995 to 1996, he was a Visiting Scholar with the University of Sheffield, Sheffield, U.K. His research interests include image processing, system identification and control, neural networks, and cognitive science.
\end{IEEEbiography}
\vspace{-15pt}
\begin{IEEEbiography}[{\includegraphics[width=1in,height=1.25in,clip,keepaspectratio]{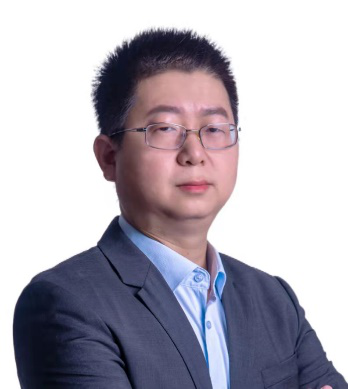}}]{Huimin Lu}
	received the B.E. degree in automation, M.E. and Ph.D. degrees in control science and engineering from the National University of Defense Technology (NUDT), Changsha, China, in 2003, 2005, and 2010, respectively. He joined the College of Intelligence Science and Technology, NUDT, in 2010, where he is now a Professor. His research interests include mobile robotics, mainly robot vision, multi-robot coordination, and human-robot interaction.
\end{IEEEbiography}

\vfill

\end{document}